\documentclass{article}
\usepackage{etoolbox}
\usepackage[preprint]{corl_2026}   %
\usepackage{booktabs}
\usepackage{comment}
\usepackage{graphicx}
\usepackage{algorithm}
\usepackage{algpseudocode}
\usepackage{amsmath}
\usepackage{amssymb}
\usepackage{xcolor}
\definecolor{linkpink}{HTML}{D6336C}
\usepackage{multirow}
\usepackage{subcaption}
\usepackage{url}
\usepackage{kotex}
\usepackage{xspace}
\usepackage{enumitem}
\usepackage{tikz}
\usepackage{silence}
\usetikzlibrary{positioning,arrows.meta,shapes.geometric,calc,fit,backgrounds,decorations.pathreplacing,patterns}
\newcommand{\method}{BinTrack\xspace}
\newcommand{\dataset}{GangnamLoop\xspace}
\newcommand{\bs}{\textsc{BinSearch}\xspace}

\newcommand{\red}[1]{\textcolor{red}{#1}}
\definecolor{teampurple}{RGB}{140,40,170}

\usepackage{titlesec}                                                    %
\titlespacing*{\section}{0pt}{8pt plus 2pt minus 2pt}{4pt plus 1pt minus 1pt}      %
\titlespacing*{\subsection}{0pt}{6pt plus 2pt minus 2pt}{3pt plus 1pt minus 1pt}   %
\titlespacing*{\paragraph}{0pt}{4pt plus 1pt minus 1pt}{1em}                       %
\newtoggle{draftmode}
\toggletrue{draftmode} 
\newif\ifshowbody
\newif\ifshowappendix
\showbodytrue
\showappendixtrue
\definecolor{authorA}{RGB}{200,30,30}
\definecolor{authorB}{RGB}{30,100,200}
\newcommand{\makeauthor}[2]{%
  \expandafter\newcommand\csname #1\endcsname[1]{%
    \iftoggle{draftmode}{\textcolor{#2}{##1}}{##1}%
  }%
  \expandafter\newcommand\csname #1note\endcsname[1]{%
    \iftoggle{draftmode}{\textcolor{#2}{\textbf{[#1: ##1]}}}{}%
  }%
}
\makeauthor{dy}{authorA}
\makeauthor{cw}{authorB}
\title{Binary Tracking for Spatial QA and Navigation\\
       with Open Vision-Language Models}
\author{%
  \end{tabular}%
  \begin{minipage}{5.5in}\centering
    \mbox{\bf Dongbin Na\thanks{Equal contribution.}\thanks{Correspondence to \texttt{dongbinna@postech.ac.kr} and \texttt{dooyoung@rgarobot.com}.}\thanks{The project page: \textcolor{linkpink}{\url{https://ndb796.github.io/BinaryTracking}}}\quad Chanwoo Kim\footnotemark[1]\quad Soonbin Rho
          \quad Giyun Choi \quad Gangbok Lee \quad Dooyoung Hong\footnotemark[2]}\\
    RGA Inc.\\
    \texttt{\{dongbinna, cwkim, hasbro425, cky, gb.lee, dooyoung\}@rgarobot.com}%
  \end{minipage}%
  \begin{tabular}[t]{c}%
}
\begin{document}
\ifshowbody                 %
\maketitle
\begin{abstract}
This work addresses spatial question answering for service robots traversing long egocentric routes.
Given a query such as \textit{``where can I find a dry cleaner on the way back home?''}, the system returns a metric coordinate that downstream navigation components can act on.
Prior Spatial Question Answering approaches leverage retrieval-augmented agents built on closed-source models such as GPT-4o for path exploration. 
However, robots operating in the real world often cannot reliably depend on online closed-source models due to network instability, communication latency, and deployment cost.
It creates a need for open-source based Spatial Question Answering approaches that can run onboard the robot, yet prior research in this direction remains limited.
This work proposes \textit{BinTrack}, a simple yet effective, fully open-source spatial-localization agent that leverages the temporal ordering of a robot's trajectory. 
BinTrack performs a binary search over the trajectory segments between two anchor landmarks identified from a query.
It improves overall accuracy by up to $22.8$\% over other open-source implementations and even matches the reported closed-source model result on the global category of the SpaceLocQA benchmark, the most challenging setting that has so far required strong reasoning agents such as GPT-4o.
Furthermore, its optimized inference strategy consistently yields more than a $1.5\times$ inference speedup over previous approaches.
Finally, this work releases \dataset, a novel and practical multi-trip outdoor benchmark collected by deploying a real quadruped robot on public streets with the anonymization policy.
It revisits the same locations under different outdoor conditions and pairs the robot's low viewpoint with the human owner's.
% The source codes and datasets are publicly available at \textcolor{linkpink}{\url{https://github.com/ndb796/BinaryTracking}}.
% A (현재)
% The source codes and datasets are publicly available at \textcolor{linkpink}{\url{https://github.com/ndb796/BinaryTracking}}.
% B (수정)
All experiments use only open-source models, ensuring full reproducibility.
The source codes and datasets are publicly available at \textcolor{linkpink}{\url{https://github.com/ndb796/BinaryTracking}}.
\end{abstract}
\keywords{Spatial Question Answering; Robot Memory; Vision-Language Models; Open-Source Reproducibility, Robot Navigation}
\section{Introduction}
\label{sec:intro}
Spatial Question Answering (SQA) enables a robot to construct a searchable memory of observed entities and their locations from long-term egocentric video and odometry, and to infer query-relevant coordinates by jointly considering semantic and spatial constraints.
Given spatial queries such as \textit{``Where did I see the dry cleaner this morning?''} or \textit{``Where is the nearest AED to the lakeside?''}, an SQA system outputs a target coordinate for the navigation, allowing the robot to move to the corresponding location.
The state-of-the-art (SOTA) method achieves high accuracy on basic and local queries~\cite{mao2025metamemory}, but relies on a closed-source model (i.e., GPT-4o) as both the planning agent and the visual verifier, as do earlier retrieval-augmented SQA systems~\cite{anwar2024remembr,xie2024embodiedrag}.
This dependency is incompatible with the navigation use case in many real-world settings. 
A robot guiding a human to the predicted coordinate may not assume reliable internet connectivity and tolerate the per-step latency of round-tripping every retrieval to a closed-source model.
Moreover, repeatedly sending video observations and queries to such a model incurs substantial operational cost.
A natural remedy is to deploy open-source models of comparable size onboard the robot, but this substitution causes existing frameworks to degrade sharply~\cite{yang2024qwen25,bai2025qwen25vl}, especially on global queries that require multi-step reasoning along the trajectory between two landmarks (Appendix~A).
This performance gap stems not from the limited reasoning capability of open-source models but from the lack of an algorithmic primitive that matches the structure of the SQA data.
In particular, global queries require candidate verification and multi-step reasoning along the route, but prior frameworks rely on closed-source models to perform these key query-time steps.
This work instead exploits the structure already present in the robot trajectory.
A robot's trajectory forms a temporally ordered sequence of spatially indexed observations, and the two anchor landmarks named by a query define a contiguous subsequence.
This structure reduces target localization within the route to a one-dimensional search problem.
This work proposes \textit{BinTrack}, a fully open-source spatial-localization agent built on Binary Tracking, a structured retrieval primitive that recursively halves the trajectory interval between two anchors, ranks candidates within each half by semantic similarity, and descends into the half with stronger evidence.
\textit{BinTrack} pairs this primitive with a multi-view memory that captures each segment from overview, center (focal), and detail perspectives, compensating for the limitations of smaller open-source verifiers without any closed-source model.
Despite its simplicity, \textit{BinTrack} outperforms other open-source implementations by up to $22.8$\,\% and even surpasses the best reported closed-source model result on the SpaceLocQA benchmark.
It also achieves more than a $1.5\times$ inference speedup over previous baseline approaches.
Finally, this work releases \dataset, a multi-trip outdoor benchmark recorded by a real quadruped robot on public urban streets, which addresses limitations of existing SQA benchmarks~\cite{mao2025metamemory, anwar2024remembr} and is detailed in Section~\ref{sec:dataset}.
\paragraph{Technical contributions:}
\begin{enumerate}[itemsep=0pt,topsep=2pt]
    \item This work introduces \textit{BinTrack}, a fully open-source SQA agent that uses Binary Tracking as its core retrieval primitive to exploit the temporal ordering of robot trajectories.
    \item A fair open-source evaluation shows that \textit{BinTrack} outperforms implementations of the two leading prior systems across all categories of the SpaceLocQA benchmark, achieving up to a $22.8$\,\% improvement over open-source baselines.
    \item It also surpasses the reported closed-source model result on the SpaceLocQA benchmark, while using only open-source components.
    \item \dataset provides a multi-trip outdoor SQA benchmark for evaluating revisit handling and cross-domain robustness in service robot deployments.
\end{enumerate}
\section{Related Work}
\label{sec:related}
\paragraph{Spatial question answering for embodied agents.}
Previous studies cast robot memory as a queryable database of egocentric observations. 
ReMEmbR introduces the NaVQA benchmark and a retrieval-augmented agent that issues function calls over a vector database of caption embeddings produced by a Visual-Language Model (VLM)~\cite{anwar2024remembr}.
Embodied-RAG organizes the same type of memory as a hierarchical semantic forest, enabling kilometer-scale retrieval~\cite{xie2024embodiedrag}.
Meta-Memory extends this line of work by augmenting caption-only memory with raw images that the verifier can directly inspect~\cite{mao2025metamemory} .
Its retrieval process uses three orchestrated tools, including semantic-similarity retrieval, spatial-range retrieval, and a graph-based memory-integration tool that constructs a waypoint graph at query time~\cite{dijkstra1959note}.
Both methods obtain their headline results using a closed-source model as the agent and verifier. 
However, simply replacing the closed-source models used in these methods with comparably sized open-source models causes their accuracy to drop sharply on global queries.
\paragraph{Memory representations for robots.}
Other approaches compress the environment into 3D scene graphs or open-vocabulary semantic maps that excel at object-centric lookups over a fixed set of nodes or grid cells~\cite{hughes2022hydra,gu2024conceptgraphs,werby2024hovsg,huang2022vlmaps,shafiullah2022clipfields,peng2023openscene}. However, this compression discards two cues essential for multi-hop spatial reasoning, the temporal order of observations and the intermediate frames between landmarks, so queries such as \textit{``What lies on the way from A to B?''} can no longer be answered, since the notion of ``between'' is lost.
\paragraph{Embodied and episodic question answering.}
Embodied QA requires an agent to navigate during inference~\cite{das2018eqa, majumdar2024openeqa}, and episodic memory QA retrieves relevant moments from a stored observation sequence~\cite{datta2022emqa}.
SQA shares the use of accumulated egocentric experience but differs in that the answer must be a metric coordinate rather than a category label, and spatial relations such as ``near,'' ``between,'' and ``on the way from'' must be operationalized for navigation.
\paragraph{Cognitive maps and long-video agents.}
Yang et al. show that explicitly generating a cognitive map improves multi-modal LLM spatial reasoning~\cite{yang2025thinkinginspace,tolman1948cogmaps}.
Meta-Memory adopts a related idea through its memory-integration tool, which constructs a waypoint graph at query time and performs graph search over retrieved memory~\cite{mao2025metamemory}.
Both approaches introduce explicit graph structures to support spatial reasoning.
In contrast, long-video agents handle temporal question categories, but they do not ground their answers in metric coordinates required for robot navigation~\cite{wang2024videoagent,fan2024videoagent2}.
\section{Problem Definition}
\label{sec:problem}
\paragraph{Spatial Question Answering.}
An SQA instance consists of a robot trajectory $T=\{(I_t, \mathbf{x}_t)\}_{t=1}^{n}$ and a natural-language query $Q$, where $I_t$ denotes the egocentric image observed at time $t$ and $\mathbf{x}_t=(x_t,y_t)$ denotes the corresponding 2D robot coordinate. 
Following prior work, the system first converts the raw trajectory into a searchable memory~\cite{mao2025metamemory,anwar2024remembr}. 
The trajectory is divided into $\Delta t$-second segments $\mathcal{S}=\{S_1,\ldots,S_N\}$, and each segment $S_i$ contains a subset of temporally adjacent frames.
The system samples several evenly spaced frames from each segment, summarizes their visual content with a VLM, and embeds the resulting captions with a text encoder. 
Each memory entry stores the caption embedding, the segment identifier, and a representative pose $\mathbf{p}_i=(x_i,y_i)$, computed from the poses of frames in the segment $S_i$.
The memory system indexes these entries in a vector database so that semantic retrieval can retrieve candidate segments $S_i$ for downstream spatial reasoning.
Given $(T,Q)$, the system parses the query into semantic targets and, when available, spatial constraints such as nearby objects, anchor landmarks, or route constraints.
It then retrieves candidate segments $S_i$ from the vector database using the query embedding and applies spatial filtering or structured search according to the query type.
For a basic query, retrieval mainly searches for segments $S_i$ whose captions match the target object.
For a local query, the system combines semantic evidence from multiple nearby objects within a local region.
For a global query, the system must reason over a contiguous interval of segment indices defined by anchor landmarks or route constraints.
The retrieved candidates are then verified using visual evidence, and the system outputs the representative pose $\mathbf{p}_i$ associated with the selected segment $S_i$ as a 2D metric coordinate $\hat{\mathbf{p}}\in\mathbb{R}^2$.
The prediction is considered correct if $\lVert\hat{\mathbf{p}}-\mathbf{p}^{\star}\rVert_2 < \tau$, where $\mathbf{p}^{\star}$ is the ground-truth coordinate and $\tau=15\mathrm{m}$ is the default distance threshold following the convention adopted by prior work~\cite{mao2025metamemory}.
Queries fall into three categories: \textbf{basic} queries require single-object recall, such as \textit{``Where is a dry cleaner?''}; \textbf{local} queries require integrating multiple objects within a local region, such as \textit{``Where is the trash bin next to the vending machine?''}; and \textbf{global} queries require long-horizon spatial reasoning, often under a path constraint between two landmarks, such as \textit{``Where is the convenience store on the route from my office to the subway station?''}.
Appendix~B provides implementation details, including the captioner, encoder, segment length, and embedding dimension.
\paragraph{Challenges in Global Queries.} 
A global query such as \textit{``Where is $Z$ on the route from $X$ to $Y$?''} requires the system to identify the route between two anchor landmarks and localize the target within that route.
This setting is difficult because a long trajectory often contains many semantically similar candidates for $Z$ across multiple regions.
Graph-based path search handles this route constraint by constructing a waypoint graph and reasoning over the graph across multiple steps.
Prior systems can make this strategy work with a strong closed-source model, which can coordinate candidate retrieval, route reasoning, and verifier calls throughout the multi-step process.
However, the same reasoning process becomes unreliable when the pipeline uses a smaller open-source agent.
As a result, global query performance drops sharply in the open-source deployment setting.
This motivates a simpler, structured retrieval primitive that directly exploits the temporal order already present in the robot trajectory, rather than relying on agent-driven graph reasoning.
\section{BinTrack}
\label{sec:method}
\textit{BinTrack} introduces two primary changes to existing agent-driven SQA pipelines~\citep{mao2025metamemory, anwar2024remembr}. 
It uses a multi-view memory representation and the Binary Tracking algorithm, which performs a structured search between two anchor landmarks. 
Two additional hardening mechanisms further protect the agent against ambiguous candidates, and Appendix~C describes their details.
\subsection{Multi-View Memory Representation}
\label{sec:memory}
Prior work represents each segment with a single caption generated from a $2{\times}2$ concatenation of four evenly spaced frames~\cite{mao2025metamemory}.
This representation becomes unreliable with 7B-class open-source VLMs~\cite{bai2025qwen25vl} because important objects may occupy only a small region of the combined image, and a single caption must summarize multiple viewpoints, often producing generic descriptions such as \textit{``a street with shops''} rather than discriminative details such as \textit{``a coffee shop named BANAPRESSO on the right.''} 
The proposed method instead represents each segment with three complementary views from the same open-source VLM under different prompts~\cite{bai2025qwen25vl}.
The \emph{full} view captures the entire scene from the $2{\times}2$ concatenation, the \emph{center} view describes objects directly ahead of the robot in the central frame, and the \emph{detail} view captures visible storefronts, signs, and readable text. 
The memory system embeds the three captions with the same open-source text encoder and stores each caption as a separate entry with a view identifier in a local vector database~\cite{lee2024mxbai, wang2021milvus}.
Semantic retrieval ranks candidates across all views jointly.
Appendix~D lists the per-view prompts and component identifiers.
\subsection{Binary Tracking}
\label{sec:bs}
Existing SQA systems either rely on simple semantic retrieval or perform graph-based path search after re-approximating the already available robot trajectory as a waypoint graph.
However, a robot trajectory naturally forms a sequence of segments ordered by recording time. 
The two anchors in a global query define a contiguous segment interval $[i_X, i_Y]$, and the answer must lie within this interval if the robot follows a continuous route between the anchors.
This observation formulates the search for $Z$ within the interval as a one-dimensional search problem, making binary search a natural and efficient procedure for exploiting this structure.
\begin{algorithm}[ht]
\caption{Binary tracking: search over ordered segment indices between two anchors.}
\label{alg}
\small
\footnotesize
\begin{algorithmic}[1]
\Function{\bs}{$Z, \mathbf{p}_X, \mathbf{p}_Y; k{\text{leaf}}, k$}
\State $i_X \gets \arg\min_i \lVert\mathbf{p}_i - \mathbf{p}_X\rVert_2$;\quad
$i_Y \gets \arg\min_i \lVert\mathbf{p}_i - \mathbf{p}_Y\rVert_2$
\Comment{anchor segment indices}
\State $(\ell, r) \gets (\min(i_X, i_Y)+1, \max(i_X, i_Y)-1)$
\While{$r - \ell > k_{\text{leaf}}$}
\State $m \gets \lfloor (\ell + r)/2 \rfloor$
\State $L, R \gets$ top-$k$ semantic scores for $Z$ on $[\ell, m]$, $[m{+}1, r]$
\If{$\mathrm{evid}(L) \geq \mathrm{evid}(R)$}
\Comment{pick the half with stronger evidence}
\State $r \gets m$
\Else
\State $\ell \gets m + 1$
\EndIf
\EndWhile
\State $\hat{i} \gets$ \Call{Verify}{top-$k$ on $[\ell, r]$, $Z$}
\State \Return $\mathbf{p}_{\hat{i}}$
\EndFunction
\end{algorithmic}
\end{algorithm}
Algorithm~\ref{alg} summarizes the Binary Tracking procedure.
The algorithm takes a parsed route-constrained query and searches the trajectory interval between two anchor landmarks.
Instead of constructing a waypoint graph, it repeatedly compares the semantic evidence in the left and right halves of the current interval and keeps the more promising half.
Let $Z$ denote the target entity in the query, and let $X$ and $Y$ denote the two anchor landmarks that define the route constraint.
The query parser returns the target $Z$, and metric anchor coordinates $\mathbf{p}_X,\mathbf{p}_Y \in \mathbb{R}^2$.
For each trajectory segment $S_i$, let $\mathbf{p}_i \in \mathbb{R}^2$ denote its representative coordinates.
Binary Tracking first locates the anchor segments $i_X = \arg\min_i \|\mathbf{p}_i - \mathbf{p}_X\|_2$ and $i_Y = \arg\min_i \|\mathbf{p}_i - \mathbf{p}_Y\|_2$, and initializes the search interval $(\ell,r)$ as the contiguous segment range between the two anchors.
During search, $m=\lfloor(\ell+r)/2\rfloor$ denotes the midpoint of the current interval.
The method computes evidence scores for the left interval $[\ell,m]$
and the right interval $[m+1,r]$,
where the evidence score for an interval $\mathrm{evid}(\cdot)=\alpha\cdot\mathrm{mean}_{i\in H}\sigma(i)+\beta\cdot\max_{i\in H}\sigma(i)$.
The algorithm keeps the half with the larger evidence score.
The loop stops when the interval length becomes no larger than the leaf size $k_{\text{leaf}}$.
The verifier then selects the final segment $\hat{i}$ from the top-$k$ candidates in the leaf interval, and the system returns $\mathbf{p}_{\hat{i}}$ as the predicted coordinate.
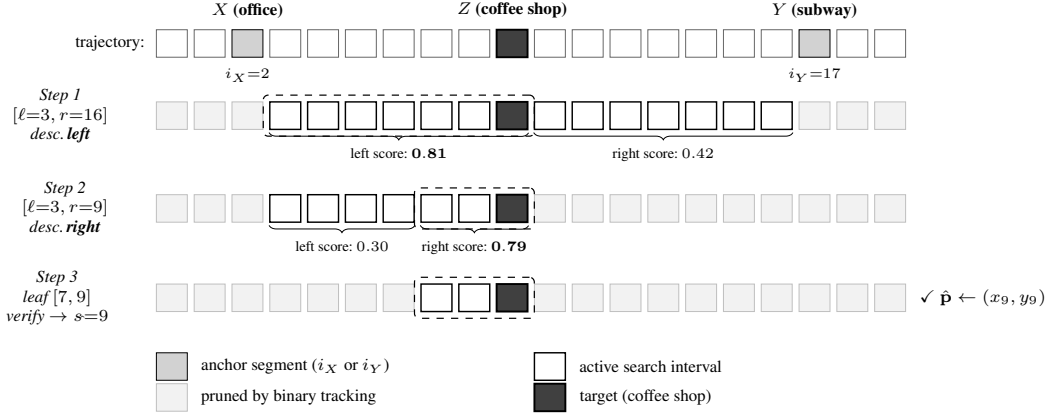
\begin{figure}[ht]
  \centering
  \resizebox{1.00\linewidth}{!}{%
  \begin{tikzpicture}[font=\footnotesize,
                       seg/.style={draw, rounded corners=0pt,
                                   minimum width=4.5mm, minimum height=4mm,
                                   inner sep=0pt, line width=0.4pt},
                       anchor_seg/.style={seg, fill=gray!35, draw=black},
                       target_seg/.style={seg, fill=black!75, draw=black, text=white, line width=0.8pt},
                       active_seg/.style={seg, fill=white, draw=black, line width=0.6pt},
                       pruned_seg/.style={seg, fill=gray!10, draw=gray!50,
                                          text=gray!50},
                       lbl/.style={font=\scriptsize\itshape, align=center},
                       brc/.style={decorate, decoration={brace, amplitude=3pt,
                                                          mirror, raise=1pt}, line width=0.4pt},
                       win/.style={draw=black, dashed, rounded corners=2pt, line width=0.5pt,
                                   inner sep=2.4pt}]
    \def\nseg{20}
    \foreach \i in {0,...,19} {
      \pgfmathsetmacro{\xpos}{\i*0.55}
      \pgfmathtruncatemacro{\isAnchorX}{equal(\i,2)}
      \pgfmathtruncatemacro{\isAnchorY}{equal(\i,17)}
      \pgfmathtruncatemacro{\isTarget}{equal(\i,9)}
      \ifnum\isAnchorX=1
        \node[anchor_seg] (s\i) at (\xpos,0) {};
      \else\ifnum\isAnchorY=1
        \node[anchor_seg] (s\i) at (\xpos,0) {};
      \else\ifnum\isTarget=1
        \node[target_seg] (s\i) at (\xpos,0) {};
      \else
        \node[seg, fill=white, draw=black!60] (s\i) at (\xpos,0) {};
      \fi\fi\fi
    }
    \node[above=1pt of s2,  font=\scriptsize\bfseries]   {$X$ (office)};
    \node[above=1pt of s9,  font=\scriptsize\bfseries]   {$Z$ (coffee shop)};
    \node[above=1pt of s17, font=\scriptsize\bfseries]   {$Y$ (subway)};
    \node[left=1pt of s0, font=\scriptsize] {trajectory:};
    \node[below=1pt of s2,  font=\tiny]  {$i_X{=}2$};
    \node[below=1pt of s17, font=\tiny]  {$i_Y{=}17$};
    \node[anchor=east, lbl] at (-0.8, -1.05)
         {Step 1\\$[\ell{=}3, r{=}16]$\\desc.\,\textbf{left}};
    \begin{scope}[yshift=-1.05cm]
      \foreach \i in {0,...,19} {
        \pgfmathsetmacro{\xpos}{\i*0.55}
        \pgfmathtruncatemacro{\inWin}{\i>2 && \i<17}
        \pgfmathtruncatemacro{\isTarget}{equal(\i,9)}
        \pgfmathtruncatemacro{\inLeftHalf}{\i>2 && \i<10}
        \pgfmathtruncatemacro{\inRightHalf}{\i>9 && \i<17}
        \ifnum\isTarget=1
          \node[target_seg] (a\i) at (\xpos,0) {};
        \else
          \ifnum\inLeftHalf=1
            \node[active_seg] (a\i) at (\xpos,0) {};
          \else\ifnum\inRightHalf=1
            \node[active_seg] (a\i) at (\xpos,0) {};
          \else
            \node[pruned_seg] (a\i) at (\xpos,0) {};
          \fi\fi
        \fi
      }
      \draw[brc] (a3.south west) -- (a9.south east)
                 node[midway, below=4pt, font=\tiny]
                 {left score: $\mathbf{0.81}$};
      \draw[brc] (a10.south west) -- (a16.south east)
                 node[midway, below=4pt, font=\tiny]
                 {right score: $0.42$};
      \node[win, fit=(a3) (a9)] {};
    \end{scope}
    \node[anchor=east, lbl] at (-0.8, -2.4)
         {Step 2\\$[\ell{=}3,r{=}9]$\\desc.\,\textbf{right}};
    \begin{scope}[yshift=-2.4cm]
      \foreach \i in {0,...,19} {
        \pgfmathsetmacro{\xpos}{\i*0.55}
        \pgfmathtruncatemacro{\isTarget}{equal(\i,9)}
        \pgfmathtruncatemacro{\inLeftHalf}{\i>2 && \i<7}
        \pgfmathtruncatemacro{\inRightHalf}{\i>6 && \i<10}
        \ifnum\isTarget=1
          \node[target_seg] (b\i) at (\xpos,0) {};
        \else
          \ifnum\inLeftHalf=1
            \node[active_seg] (b\i) at (\xpos,0) {};
          \else\ifnum\inRightHalf=1
            \node[active_seg] (b\i) at (\xpos,0) {};
          \else
            \node[pruned_seg] (b\i) at (\xpos,0) {};
          \fi\fi
        \fi
      }
      \draw[brc] (b3.south west) -- (b6.south east)
                 node[midway, below=4pt, font=\tiny]
                 {left score: $0.30$};
      \draw[brc] (b7.south west) -- (b9.south east)
                 node[midway, below=4pt, font=\tiny]
                 {right score: $\mathbf{0.79}$};
      \node[win, fit=(b7) (b9)] {};
    \end{scope}
    \node[anchor=east, lbl] at (-0.8, -3.7)
         {Step 3\\leaf $[7, 9]$\\verify $\rightarrow s_{\hat{}}{=}9$};
    \begin{scope}[yshift=-3.7cm]
      \foreach \i in {0,...,19} {
        \pgfmathsetmacro{\xpos}{\i*0.55}
        \pgfmathtruncatemacro{\isTarget}{equal(\i,9)}
        \pgfmathtruncatemacro{\inLeaf}{\i>6 && \i<10}
        \ifnum\isTarget=1
          \node[target_seg] (c\i) at (\xpos,0) {};
        \else
          \ifnum\inLeaf=1
            \node[active_seg] (c\i) at (\xpos,0) {};
          \else
            \node[pruned_seg] (c\i) at (\xpos,0) {};
          \fi
        \fi
      }
      \node[win, fit=(c7) (c9)] {};
      \node[right=2pt of c19, font=\scriptsize,
            align=left]
            {$\checkmark\ \hat{\mathbf{p}} \gets (x_9, y_9)$};
    \end{scope}
    \begin{scope}[yshift=-4.7cm, xshift=0cm]
      \node[anchor_seg] (lgX) at (0,0) {};
      \node[right=2pt of lgX, font=\scriptsize]
          {anchor segment ($i_X$ or $i_Y$)};
      \node[active_seg] (lgA) at (5.5,0) {};
      \node[right=2pt of lgA, font=\scriptsize]
          {active search interval};
      \node[pruned_seg] (lgP) at (0,-0.45) {};
      \node[right=2pt of lgP, font=\scriptsize]
          {pruned by binary tracking};
      \node[target_seg] (lgT) at (5.5,-0.45) {};
      \node[right=2pt of lgT, font=\scriptsize]
          {target (coffee shop)};
    \end{scope}
  \end{tikzpicture}}
  \caption{Binary Tracking performs a binary search over the ordered segment indices between two anchor segments.}
  \label{fig:method-overview}
\end{figure}
Figure~\ref{fig:method-overview} illustrates the intuition behind Binary Tracking on an ordered trajectory.
The query asks for a coffee shop on the way from the office to the subway, so the two anchor segments first define the search interval between them.
Binary Tracking then repeatedly compares the left and right halves of the current interval and keeps the half with stronger evidence for the target.
In the example, the first comparison keeps the left half, and the second comparison keeps the right half, reducing the search interval to a small leaf interval.
A verifier then selects the final target segment from the remaining candidates, and the system returns the pose of that segment as the predicted coordinate.
Let $n$ denote the number of segments between the two anchors, and let $c_{\mathrm{ret}}$ and $c_{\mathrm{ver}}$ denote the costs of one semantic retrieval and one verifier call, respectively.
Binary Tracking repeatedly halves the search interval until the remaining interval contains at most $k_{\mathrm{leaf}}$ segments.
This requires $\lceil \log_2(n/k_{\mathrm{leaf}}) \rceil$ iterations.
At each iteration, the method performs one semantic retrieval on the left half and one on the right half, and it calls the verifier only once after reaching the final leaf interval.
Thus, the total cost is 
\begin{equation}
C_{\mathrm{BT}}(n) = 2c_{\mathrm{ret}}\lceil \log_2(n/k_{\mathrm{leaf}}) \rceil + c_{\mathrm{ver}}
= O(\log n)\cdot c_{\text{ret}} + c_{\text{ver}},
\end{equation}
which scales logarithmically with the interval length.
The naive worst case where the verifier inspects all $n$ candidates costs $c_{\mathrm{ret}} + n c_{\mathrm{ver}}$. Binary Tracking instead invokes the expensive verifier only on the final leaf, which matters because $c_{\mathrm{ver}} \gg c_{\mathrm{ret}}$ for 7B-class open-source models.
\subsection{Agent Loop and Hardening Mechanisms}
\label{sec:agent}
\textit{BinTrack} uses an open-source instruction-tuned language model as the planning agent~\cite{yang2024qwen25,lin2024awq}.
The agent runs a thought-and-action loop, where each step selects one action from a small set covering semantic and spatial retrieval primitives, the proposed Binary Tracking, and the final coordinate output. 
Appendix~E lists the full tool set, tool descriptions, and per-trajectory filtering applied at each retrieval step.
This simple design, driven primarily by Binary Tracking and multi-view memory, substantially improves accuracy under the open-source constraint.
Two additional lightweight hardening mechanisms further improve robustness when the verifier uses a smaller open-source VLM.
The anchor-aware verification pool balances verifier candidates between semantically similar matches and anchor-consistent matches.
The lock-on-Y rerank reorders candidates using a secondary anchor extracted from the query and provides a fallback when the agent does not converge within its step budget.
Appendix~F details both mechanisms.
\section{The \dataset Benchmark}
\label{sec:dataset}
\vspace{-3mm}
\begin{figure}[h]
\centering
\includegraphics[width=1.00\textwidth]{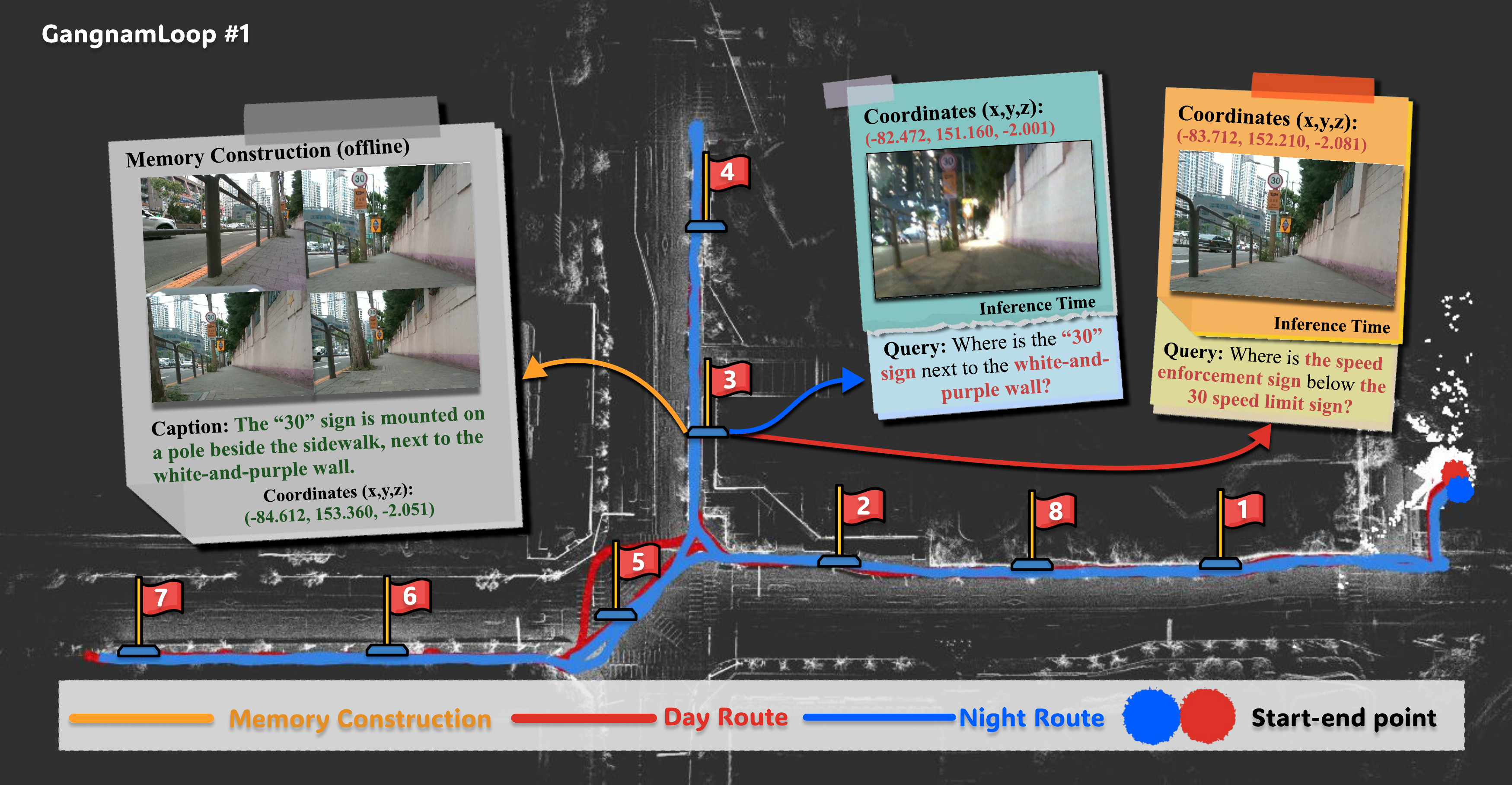}
\caption{Overview of \dataset on a representative round-trip route (office $E$ to $B$ and back). Day (red) and night (blue) trajectories are aligned on a common SLAM map.}
\label{fig:gangnam-overview}
\end{figure}
Public SQA datasets typically consist of single one-pass tours collected in indoor or campus environments~\cite {mao2025metamemory, anwar2024remembr}. 
However, real service robots repeatedly travel the same routes and encounter appearance changes caused by illumination, weather, and foreground objects over time.
They also face a viewpoint mismatch, as a quadruped service robot records video from a low viewpoint under approximately $0.8\,\mathrm{m}$, while a human user typically asks spatial questions from an eye-level viewpoint in roughly the $1.5$--$1.9\,\mathrm{m}$ range.
Existing datasets do not sufficiently capture revisits, domain shifts, and viewpoint mismatch in real service-robot deployments.
To address this gap, this work releases \dataset, a multi-trip outdoor SQA benchmark recorded by a real quadruped robot on public urban streets. 
Figure~\ref{fig:gangnam-overview} gives an overview of one representative round-trip route, showing the day and night trajectories on the SLAM map together with the memory-construction view examples and a test query example used throughout the benchmark.
\paragraph{Recording platform.}
\dataset uses a Unitree Go2~\cite{unitree2024go2} with an Intel RealSense depth camera D455~\cite{realsense2020D455}, a Livox MID-360 LiDAR~\cite{livox2024mid360} with a built-in IMU.
An onboard LiDAR-Inertial SLAM pipeline inspired by Point-LIO estimates the per-frame poses~\cite{he2023pointlio}.
A human user walks alongside the robot during collection, and a subset additionally captures the user's first-person view through a head-mounted camera. 
\dataset reflects how real service robots operate in practice, unlike public SQA benchmarks that use single-pass tours.
A real quadruped robot records the benchmark data on public urban streets while revisiting the same locations under paired day and night conditions.
The benchmark provides dual viewpoints from a low camera mounted on the robot and a head-mounted camera worn by the human user.
It also includes anonymized real pedestrian scenes that rarely appear in existing SQA datasets.
A direct comparison with related egocentric and embodied benchmarks is included in Appendix~E.
The full benchmark dataset contains 8 round-trip recordings. 
Each round trip starts and ends at a fixed office location ($E$) and visits one of four nearby subway stations ($A$--$D$) along the route.
The collection schedule records each destination once during the day and once at night.
Table~\ref{tab:gangnam-schedule} summarizes the resulting four pairs of day and night and the recording statistics.
Appendix~F lists the mapping between the symbols $E$, $A$, $B$, $C$, $D$, and the actual identities of the office and the station, together with the full collection schedule.
  \begin{table}[ht]
    \centering\footnotesize
    \setlength{\tabcolsep}{4pt}
    \renewcommand{\arraystretch}{0.9}
    \caption{\dataset recording schedule.}
    \label{tab:gangnam-schedule}
    \begin{tabular}{cllllrrr}
      \toprule
      \# & Day & Condition & Destination (round trip) & Pair w/ &  Duration & RGB frames & Queries \\
      \midrule
      1 & Day~1 & day   & $E\,\to\,A\,\to\,E$    & \#8 & 14~m~43~s & 25{,}545 & 45 \\
      2 & Day~1 & day   & $E\,\to\,B\,\to\,E$     & \#7 & 12~m~37~s & 22{,}020 & 45 \\
      3 & Day~1 & night & $E\,\to\,C\,\to\,E$    & \#6 & 32~m~47~s & 56{,}883 & 45 \\
      4 & Day~1 & night & $E\,\to\,D\,\to\,E$    & \#5 & 45~m~12~s & 78{,}507 & 45 \\
      5 & Day~2 & day   & $E\,\to\,D\,\to\,E$    & \#4 & 49~m~40~s & 86{,}208 & 45 \\
      6 & Day~2 & day   & $E\,\to\,C\,\to\,E$    & \#3 & 35~m~19~s & 61{,}397 & 45 \\
      7 & Day~2 & night & $E\,\to\,B\,\to\,E$     & \#2 & 13~m~41~s & 23{,}753 & 45 \\
      8 & Day~2 & night & $E\,\to\,A\,\to\,E$    & \#1 & 17~m~~~3~s & 29{,}487 & 45 \\
      \midrule
      \multicolumn{5}{r}{\textbf{Totals (committed in open release):}} & 221~m~~~1~s & 383{,}800 & \textbf{360} \\
      \bottomrule
    \end{tabular}
  \end{table}
\section{Experiments}
\label{sec:exp}
All experiments can be run on a single workstation with two NVIDIA RTX 6000 Ada Generation GPUs ($48\,\mathrm{GB}$ each) and no external API access, using only open-source models for the captioner, text encoder, planning agent, and visual verifier. 
The whole pipeline also fits within $20\,\mathrm{GB}$ of VRAM with 7B-class counterparts. 
Full model identifiers, prompts, and hyperparameters are in Appendix~D. 
The evaluation uses the 270-query SpaceLocQA benchmark ($90$ basic, $90$ local, $90$ global across 6 trajectories) and all 8 \dataset recordings.
The main metric is success rate under a metric tolerance $\tau$, with $\tau=15\,\mathrm{m}$ as the headline setting.
\subsection{Experiment results}
\label{sec:exp-main}
\begin{table}[h!]
  \centering\small
\footnotesize
  \red{\renewcommand{\arraystretch}{0.8}}
  \caption{SQA comparison results on SpaceLocQA (270 queries, success at $\tau=15\,\mathrm{m}$). $\Delta_{\text{global}}$ denotes the global-category change from the closed-source model backbone to its comparably sized open-source counterpart.}
  \label{tab:main}
  \begin{tabular}{llccccc}
    \toprule
    Method & Backbone & Basic ↑ & Local ↑ & Global ↑ & Overall ↑ & $\Delta_{\text{global}}$ \\
    \midrule
    \multirow{2}{*}{Meta-Memory~\citep{mao2025metamemory}}
      & closed-source & 67.8 & 61.8 & 62.2 & 63.9 & \multirow{2}{*}{$-29.6$} \\
      & open-source$^\dagger$ & 50.0 & 51.1 & 32.6 & 44.6 & \\
    \midrule
    \multirow{2}{*}{ReMEmbR~\citep{anwar2024remembr}}
      & closed-source & 58.5 & 57.8 & 46.3 & 54.2 & \multirow{2}{*}{$-13.7$} \\
      & open-source$^\dagger$ & 56.7 & 61.1 & 32.6 & 50.1 & \\
    \midrule
    \textbf{\method (ours)} & open-source & \textbf{74.4} & \textbf{65.6} & \textbf{62.2} & \textbf{67.4} & --- \\
    \bottomrule
  \end{tabular}
  \\[2pt]
  {\footnotesize $^\dagger$ Reproduced by us under the fair constraint (Qwen2.5-32B-AWQ agent and Qwen2.5-7B verifier).}
\end{table}
Table~\ref{tab:main} reports success rates in $\tau{=}15\,\mathrm{m}$ on SpaceLocQA for both the closed-source model and the open-source backbone of prior systems. 
Under the fair open-source constraint, \method outperforms both open-source baselines in every category, with the largest gain on global queries ($+29.6\,\%$ over open-source Meta-Memory). 
More importantly, \method also \emph{surpasses} the best published closed-source model result of Meta-Memory by $+6.6$, $+3.8$, $0.0$, and $+3.5\,\%$ on basic, local, global, and overall.
The result on the global category is particularly noteworthy. Prior work has shown that global queries are the hardest setting and have so far been solvable only with strong reasoning agents such as GPT-4o~\cite{mao2025metamemory}, yet \method matches that closed-source result ($62.2\,\%$) using only open-source local models, narrowing a gap that the open-source substitution alone widens by $-29.6\,\%$.
\begin{table}[h]
  \centering\footnotesize
  \setlength{\tabcolsep}{4pt}
  \renewcommand{\arraystretch}{1.0}
  \caption{Per-sequence and per-category comparison on SpaceLocQA (success at $\tau=15\,\mathrm{m}$). The three baselines are the closed-source model results from the Meta-Memory paper.}
  \label{tab:per-seq-cat}
  \begin{tabular}{llccccccc}
    \toprule
    Category & Method & Seq 0 & Seq 1 & Seq 2 & Seq 3 & Seq 4 & Seq 5 & Avg ↑ \\
    \midrule
    \multirow{4}{*}{Basic}
      & ReMEmbR~\cite{anwar2024remembr}        & 46.7 & 53.3 & \textit{82.2} & 71.1 & 40.0 & 57.8 & 58.5 \\
      & Embodied-RAG~\cite{xie2024embodiedrag} & 42.2 & 37.8 & 71.1 & 62.2 & 35.6 & 68.9 & 53.0 \\
      & Meta-Memory~\cite{mao2025metamemory}   & \textit{53.3} & \textit{57.8} & \textbf{84.4} & \textit{73.3} & \textit{57.8} & \textit{80.0} & \textit{67.8} \\
      & \method (ours)                         & \textbf{66.7} & \textbf{60.0} & 73.3 & \textbf{86.7} & \textbf{73.3} & \textbf{86.7} & \textbf{74.4} \\
    \midrule
    \multirow{4}{*}{Local}
      & ReMEmbR~\cite{anwar2024remembr}        & \textit{68.9} & \textbf{60.0} & 51.1 & \textbf{66.7} & 44.4 & 55.6 & 57.8 \\
      & Embodied-RAG~\cite{xie2024embodiedrag} & 62.2 & \textit{48.9} & \textit{57.8} & 53.3 & \textit{46.7} & 62.2 & 55.2 \\
      & Meta-Memory~\cite{mao2025metamemory}   & 66.7 & \textbf{60.0} & 53.3 & \textit{64.4} & \textbf{60.0} & \textit{66.7} & \textit{61.8} \\
      & \method (ours)                         & \textbf{80.0} & 46.7 & \textbf{73.3} & 60.0 & \textbf{60.0} & \textbf{73.3} & \textbf{65.6} \\
    \midrule
    \multirow{4}{*}{Global}
      & ReMEmbR~\cite{anwar2024remembr}        & 51.1 & 44.4 & 60.0 & 37.8 & 51.1 & 33.3 & \textit{46.3} \\
      & Embodied-RAG~\cite{xie2024embodiedrag} & 37.8 & 42.2 & 40.0 & 20.0 & 40.0 & 44.4 & 37.4 \\
      & Meta-Memory~\cite{mao2025metamemory}   & \textit{60.0} & \textit{62.2} & \textbf{82.2} & \textit{55.6} & \textbf{60.0} & \textbf{53.3} & \textbf{62.2} \\
      & \method (ours)                         & \textbf{80.0} & \textbf{66.7} & \textit{66.7} & \textbf{60.0} & \textit{53.3} & \textit{46.7} & \textbf{62.2} \\
    \bottomrule
  \end{tabular}
\end{table}
Table~\ref{tab:per-seq-cat} gives a finer-grained per-sequence comparison for each category.
The comparison includes three prior baselines, ReMEmbR, Embodied-RAG, and Meta-Memory, which use a closed-source model, GPT-4o, as reported in previous studies.
In contrast, \method uses only open-source local models.
Even under this stricter open-source setting, \method achieves the highest average accuracy in every category and ranks first on the majority of individual sequences across all six trajectories.
This consistent ranking across sequences indicates that the accuracy of \method is not driven by a few favorable trajectories but generalizes across diverse routes, which is in line with the design intent of Binary Tracking as a trajectory-structured primitive.
These results show that an open-source pipeline with Binary Tracking can match or exceed strong closed-source model baselines without any online dependency.
\subsection{Results on \dataset}
\label{sec:exp-gangnam}
The evaluation further applies \method to all 8 \dataset recordings using the same pipeline.
A key characteristic of \dataset is that it revisits the same locations multiple times under different conditions, which directly stresses the revisit-handling ability of an SQA system.
Since no closed-source model result has been reported on \dataset, the comparison focuses on the strongest open-source baseline.
\method outperforms this baseline by more than $2.5\,\%$ in overall average accuracy across the 8 recordings, confirming that Binary Tracking transfers to outdoor low-viewpoint service-robot data and to the multi-visit setting without source-code changes (full per-recording breakdown in Appendix~G).
We further expect that augmenting \dataset with additional and more challenging queries and labels will catalyze active follow-up research on stronger methods for service-robot spatial question answering.
\subsection{Ablations and Failure Analysis}
\label{sec:exp-ablations}
The ablation study examines the two core design choices of \method, Binary Tracking, and multi-view memory, by removing them individually and jointly.
The results show that both components matter.
Starting from the full model accuracy of $67.4\%$, removing Binary Tracking reduces accuracy to $63.3\%$, removing multi-view memory reduces it to $63.0\%$, and removing both further reduces it to $59.6\%$.
The degradation appears most strongly on global path-constrained queries, where Binary Tracking directly contributes to route-aware localization.
Appendix~H provides the full ablation table and detailed failure analysis.
\section{Limitations}
\label{sec:limitations}
\method, like prior systems, builds memory per trajectory. 
A naive cross-trajectory retrieval over shared landmarks gave no notable gain (Appendix~J), so principled cross-trajectory integration is left to future work. 
A more practical limitation is that the captioning stage still requires offline background processing for newly visited routes, which constrains fully online deployment.
\section{Conclusion}
\label{sec:conclusion}
\method is a fully open-source SQA agent built on Binary Tracking, a structured retrieval primitive matched to the temporal ordering of a robot's trajectory.
Using only 7B-class open-source models, \method achieves a $67.4\%$ success rate on SpaceLocQA, outperforming open-source baselines by up to $22.8$\,\% and surpassing the best reported closed-source Meta-Memory result.
This work also releases \dataset, a multi-trip outdoor SQA benchmark collected with a real quadruped service robot on public urban streets, containing 8 round-trip recordings with paired day and night conditions, dual robot and human viewpoints, and anonymized real pedestrian scenes.
Overall, the results show that trajectory-structured algorithmic primitives can narrow the gap between open-source and closed-source foundation-model pipelines for embodied spatial reasoning.
\bibliography{references}
\fi                         %
\ifshowappendix
\appendix

\newpage
\section{Closed-to-Open Substitution Gap}
\label{app:closed-open-gap}

Table~\ref{tab:closed-open-gap} quantifies the performance degradation that motivates this work.
When prior systems replace their closed-source agent and verifier with comparably sized open-source models under the fair-comparison constraint, accuracy drops sharply, especially in the \emph{global} category.
The closed-source and open-source rows match the corresponding rows in Table~\ref{tab:main}, ensuring consistency between the main results and this appendix.
This appendix additionally reports the per-category $\Delta$ values from closed-source to open-source settings for each baseline, making the category-wise concentration of the degradation explicit.
For reference, the last row reports \method, which uses only open-source components by design.

\begin{table}[ht]
    \centering\small                                    
    \caption{Effect of substituting closed-source models with comparably sized open-source models in prior SQA pipelines.}
    \label{tab:closed-open-gap}
    \begin{tabular}{llcccc}
      \toprule
      Method & Backbone & Basic ↑ & Local ↑ & Global ↑ & Overall ↑ \\
      \midrule
      \multirow{3}{*}{Meta-Memory~\citep{mao2025metamemory}}
        & closed-source & 67.8 & 61.8 & 62.2 & 63.9 \\
        & open-source$^\dagger$ & 50.0 & 51.1 & 32.6 & 44.6 \\
        & $\Delta$ (closed$\to$open) & $-17.8$ & $-10.7$ & $\mathbf{-29.6}$ & $-19.3$ \\
      \midrule
      \multirow{3}{*}{ReMEmbR~\citep{anwar2024remembr}}
        & closed-source & 58.5 & 57.8 & 46.3 & 54.2 \\
        & open-source$^\dagger$ & 56.7 & 61.1 & 32.6 & 50.1 \\
        & $\Delta$ (closed$\to$open) & $-1.8$ & $+3.3$ & $\mathbf{-13.7}$ & $-4.1$ \\
      \midrule
      \textbf{\method (ours)} & open-source & \textbf{74.4} & \textbf{65.6} & \textbf{62.2} & \textbf{67.4} \\
      \bottomrule
    \end{tabular}
    \\[2pt]
    {\footnotesize $^\dagger$ Reproduced by us under the fair
    open-source constraint.}
\end{table}

Table~\ref{tab:closed-open-gap} shows the effect of replacing the closed-source model in prior SQA pipelines with comparably sized open-source models, using a Qwen2.5-32B-AWQ agent and a Qwen2.5-7B verifier.
Unlike Table~\ref{tab:main}, this appendix reports per-category $\Delta$ values from the closed-source setting to the open-source setting for each baseline, making the category-wise degradation explicit.
The degradation concentrates in the global category for both baselines, with drops of $-29.6$ and $-13.7$ points, while the basic and local categories change much less and can even improve, as in ReMEmbR local accuracy with $+3.3$ points.

\method achieves the strongest open-source global accuracy without using any closed-source component.
The performance degradation from closed-source to open-source substitution concentrates in the \emph{global} category for both Meta-Memory ($-29.6\,\%p$) and ReMEmbR ($-13.7\,\%p$), while the basic and local categories degrade much less.
This pattern reflects the design of prior systems, where the closed-source model handles the verification and multi-step reasoning loop that global queries rely on most.
The open-source baselines drop to roughly $32.6\,\%$ global accuracy, indicating that the smaller agent cannot reliably coordinate the graph-based memory-integration pipeline.
In the same open-source setting, \method recovers $62.2\,\%$ global accuracy by replacing graph-based reasoning with the trajectory-structured Binary Tracking primitive in Section~\ref{sec:bs}.
This primitive avoids the verifier-driven multi-step loop that becomes unreliable with a 32B-class open-source agent.

\section{Implementation Details}
\label{app:implementation}

This appendix lists the captioner, encoder, segment length, embedding dimension, decoding parameters, and compute setup used throughout the experiments.
Appendix~\ref{app:prompts} lists the per-view caption prompts and component identifiers, so this appendix focuses on algorithmic and implementation hyperparameters.

\paragraph{Segmentation and captioning.}
The pipeline divides the robot trajectory into fixed-length time segments of duration $\Delta t$.
The released configuration uses non-overlapping windows with $\Delta t = 1.5\,\mathrm{s}$.
The method does not depend on this exact value, and values in the $1$--$3\,\mathrm{s}$ range show comparable behavior.
Shorter segments provide finer caption granularity but increase the number of stored segments, while longer segments reduce memory size at the cost of coarser localization.
For each segment, the pipeline samples four evenly spaced frames and arranges them into a single $2{\times}2$ image.
It then queries the video VLM three times using the three view prompts described in Appendix~\ref{app:prompts}.
The system performs captioning once per route, caches the captions on disk, and reuses the cache for all subsequent retrieval steps without invoking the captioner again.

\paragraph{Embedding and indexing.}
 The memory system embeds all three captions for each segment with
  \texttt{mxbai-embed-large-v1}~\citep{lee2024mxbai}, producing
  $1024$-dimensional embeddings.
It stores each caption as a separate entry in a local vector database~\citep{wang2021milvus}, together with its trajectory ID, segment index, and view ID.
Every retrieval call applies the per-trajectory filter described in Appendix~\ref{app:agent-bench}.

\paragraph{Decoding parameters.}
All language and vision-language model calls use deterministic decoding with temperature $0$ and top-$p$ $1$.
The agent uses a configurable step budget, set to $4$ in the released configuration, and a per-call generation cap of $400$ tokens.
This budget is sufficient for most queries because the agent usually reaches a \textsc{final} answer within four steps.
Appendix~\ref{app:detailed-results} reports the corresponding tool-call statistics.
Each retrieval call requests up to top-$k=40$ candidates, and the verifier inspects the top $12$ candidates.
Algorithm~\ref{alg} uses a leaf size of $k_{\text{leaf}}=16$.
The system exposes these values as hyperparameters, so users can choose more conservative settings, such as a larger step budget or a wider top-$k$, without changing the algorithm.

\paragraph{Compute setup.}
The experiments use a single workstation with $2 \times$ NVIDIA RTX 6000 Ada Generation GPUs, each with $48\,\mathrm{GB}$ of VRAM, and $128\,\mathrm{GB}$ of system memory.
The runtime places the agent and verifier on separate GPUs, while per-trajectory memory construction runs sequentially.
The pipeline can also replace all four model roles, including the captioner, text encoder, planning agent, and visual verifier, with 7B-class open-source counterparts.
This lightweight configuration fits within roughly $20\,\mathrm{GB}$ of VRAM on a single consumer GPU.
Appendix~\ref{app:detailed-results} provides additional details on this configuration.

\section{Hardening Mechanisms in Detail}
\label{app:hardening}

\paragraph{Anchor-aware verification pool.}
The verifier inspects a pool of $K$ candidate segments for each retrieval call.
The system constructs this pool from the top-$K$ segments ranked by caption embedding similarity to the target $Z$ when the query has no anchor.
The query may provide an anchor pose $\mathbf{p}_X$, such as the \emph{from}-pose in Binary Tracking or a pose extracted by an earlier retrieval step.
In this case, the system combines the top-$K/2$ semantic matches with the top-$K/2$ segments closest to $\mathbf{p}_X$ in Euclidean distance.
This design prevents the verifier from considering only semantically plausible candidates.
It is especially useful when the correct answer has a borderline semantic score but lies close to the relevant anchor.

\paragraph{Lock-on-Y rerank.}
The lock-on-Y rerank handles queries that ask for a target $Z$ near or closest to a secondary anchor $Y$.
This secondary anchor differs from the from and to anchors used in Binary Tracking.
The system detects $Y$ with two lightweight rules.
The first rule uses regex patterns for English expressions such as ``nearest $Z$ to $Y$'', ``$Z$ closest to $Y$'', ``$Z$ near $Y$'', and related variants.
The second rule scans the preceding sentence for known location nouns such as \emph{lakeside}, \emph{cafeteria}, \emph{entrance}, \emph{lobby}, and \emph{rooftop}.
After detecting $Y$, the agent performs a secondary retrieval for $Y$ and verifies its candidates.
The system then re-sorts the verified $Z$ candidates by Euclidean distance to the closest verified $Y$ pose.
If the agent does not converge within its step budget, the system locks the reranked top-1 candidate as a fallback.
The rerank runs only when the verifier confirms at least two distinct $Y$ poses and the reranking changes the top-1 candidate.

Both mechanisms act as agent-side wrappers.
They do not require model retraining and only reorder the candidate set inspected by the open-source verifier.
The ablation in Appendix~\ref{app:ablation-failure} shows that each mechanism provides a small but consistent gain, about $2\,\%$ overall when removed individually.
The gain appears most strongly in the global category, which supports their role as robustness layers around the verifier.

\section{Per-View Prompts and Component Identifiers}
\label{app:prompts}

\paragraph{Per-view caption prompts.}
The captioning module generates three captions for each segment of duration $\Delta t$ using the open-source video VLM described in Appendix~\ref{app:background}.
Each caption corresponds to one of the three views in Table~\ref{tab:prompts}.
The system performs captioning once per route and caches the results for subsequent retrieval.

\begin{table}[ht]
  \centering\small
  \caption{Caption prompts for the three views of each segment.}
  \label{tab:prompts}
  \begin{tabular}{lp{0.78\linewidth}}
    \toprule
    View    & Prompt template \\
    \midrule
    full    & ``You are a robot navigating an environment. The image is a $2{\times}2$ grid of 4 consecutive frames. Describe what you see across these frames: visible objects, landmarks, signs, doors, rooms, distinctive structures, and the overall scene type. Be specific and concrete. Limit to ${\sim}80$ words.'' \\
    \addlinespace
    center  & ``You are a robot. This image is a $2{\times}2$ grid of 4 close-up center crops from 4 consecutive frames. Identify ANY readable text, signs, room numbers, brand names, small objects, equipment labels, or fine details in the center of view. Be specific. Limit to ${\sim}80$ words.'' \\
    \addlinespace
    detail  & ``You are a robot. Look at the $2{\times}2$ grid of 4 consecutive frames. Now enumerate concrete nouns visible: list distinct OBJECTS, FURNITURE, EQUIPMENT, SIGNAGE, BUILDING\_FEATURES, NATURAL\_FEATURES. Use short noun phrases separated by commas, grouped by category. ${\sim}80$ words.'' \\
    \bottomrule
  \end{tabular}
\end{table}

\paragraph{Component identifiers.}
Table~\ref{tab:components} lists the exact open-source model identifiers used for each role in the pipeline, including the captioner, text encoder, planning agent, and visual verifier.
The pipeline runs all components locally and does not require any external API access during either memory construction or query-time inference.
This setup ensures that the reported results reflect a fully open-source deployment setting rather than a hybrid system that depends on commercial services.

\red{\begin{table}[ht]
  \centering\small
  \caption{Open-source component identifiers used by \method.}
  \label{tab:components}
  \setlength{\tabcolsep}{4pt}
  \begin{tabular}{@{}llp{0.40\linewidth}@{}}
    \toprule
    Role               & Identifier (HuggingFace-style)               & Notes \\
    \midrule
    Captioner          & \texttt{Qwen2.5-VL-7B-Instruct}~\citep{bai2025qwen25vl} & video-VLM, $2\times2$ grid input \\
    Visual verifier    & \texttt{Qwen2.5-VL-7B-Instruct}~\citep{bai2025qwen25vl} & same checkpoint as captioner \\
    Text encoder       & \texttt{mxbai-embed-large-v1}~\citep{lee2024mxbai} & $1024$-dim caption embeddings \\
    Planning agent     & \texttt{Qwen2.5-32B-Instruct-AWQ}~\citep{yang2024qwen25,lin2024awq} & 4-bit AWQ quantized \\
    Vector database    & Milvus~\citep{wang2021milvus} & \raggedright local, embedded, per-trajectory partitions \tabularnewline
    \bottomrule
  \end{tabular}
\end{table}}

\section{Agent Tool Set and Comparison with Related Benchmarks}
\label{app:agent-bench}

\subsection{Agent Tool Set and Per-Trajectory Filtering}
\label{app:agent-tools}

\paragraph{Tool set.}
Table~\ref{tab:tools} summarizes the tools exposed to the planning agent at each step of its thought-and-action loop.
The tool set includes retrieval primitives, Binary Tracking, and control actions for continuing reasoning or returning the final coordinate.
The set remains intentionally small so that the agent's behavior is easy to inspect and the ablation in Appendix~\ref{app:ablation-failure} can isolate the contribution of each primitive.

\begin{table}[ht]
  \centering\small
  \setlength{\tabcolsep}{4pt}
  \caption{Agent tools exposed to the planning agent at every step.}
  \label{tab:tools}
  \begin{tabular}{llp{0.55\linewidth}}
    \toprule
    Tool                                        & Symbol           & Description \\
    \midrule
    Semantic-similarity retrieval               & \textsc{SSR}    & Top-$k$ segments by cosine similarity between the query embedding and each segment's caption embedding (ranked jointly across the three views). \\
    Spatial-range retrieval                     & \textsc{SRR}    & All segments whose representative pose lies within a radius $r$ of a query point $\mathbf{p}$. \\
    Binary tracking                             & \textsc{BT}     & Structured retrieval over the ordered segment interval between two anchor landmarks; see Algorithm~\ref{alg}. (Exposed in the released code as the path-constrained retrieval action; the agent emits the action name \texttt{PATH}, which executes Binary Tracking.) \\
    Memory integration (legacy)                 & \textsc{MI}     & Graph-based path search over a waypoint graph built from retrieved landmarks~\citep{mao2025metamemory,dijkstra1959note}; kept for backward compatibility with prior pipelines. \\
    Final-answer emission                       & \textsc{final}  & Terminates the agent loop and emits a 2D pose $\hat{\mathbf{p}}$ as the predicted coordinate. \\
    \bottomrule
  \end{tabular}
\end{table}

\paragraph{Per-trajectory filtering.}
Every retrieval call, including \textsc{SSR}, \textsc{SRR}, and the internal retrieval inside \textsc{BT}, uses a per-trajectory filter that restricts results to segments from the trajectory under evaluation.
The system implements this filter as a Milvus partition tag, so the vector database applies the restriction directly without Python-side post-filtering.
This design ensures that all reported results follow the single-trajectory retrieval protocol used by SpaceLocQA and prior work~\citep{mao2025metamemory,anwar2024remembr}.
Appendix~\ref{app:cross-traj} studies a controlled relaxation of this filter.

\subsection{Comparison with Related Egocentric and Embodied Benchmarks}
\label{app:bench-comparison}

Table~\ref{tab:bench-comparison} compares \dataset with public egocentric and embodied benchmarks.
The comparison focuses on properties relevant to real service-robot SQA, including outdoor data collection, repeated visits to the same locations, paired robot and human viewpoints, and real-robot deployment.
Here, ``Repeat'' indicates multiple visits to the same location under different conditions, and ``Dual view'' indicates paired robot-view and human-view recordings of the same route.

\begin{table}[ht]
  \centering\footnotesize
  \caption{Comparison with public egocentric and embodied benchmarks.}
  \label{tab:bench-comparison}
  \begin{tabular}{lcccccc}
    \toprule
    Benchmark              & Setting        & Answer type     & Outdoor & Repeat & Dual view & Real robot \\
    \midrule
    OpenEQA~\citep{majumdar2024openeqa}      & embodied QA    & language        & --      & --     & --        & --         \\
    EQA-MP3D~\citep{das2018eqa}              & embodied QA    & language        & --      & --     & --        & --         \\
    EMQA~\citep{datta2022emqa}               & episodic QA    & category        & --      & --     & --        & --         \\
    NaVQA~\citep{anwar2024remembr}           & SQA            & coordinate      & partial & --     & --        & \checkmark \\
    SpaceLocQA~\citep{mao2025metamemory}     & SQA            & coordinate      & partial & --     & --        & --         \\
    \dataset (this work)                     & SQA            & coordinate      & \checkmark & \checkmark & \checkmark & \checkmark \\
    \bottomrule
  \end{tabular}
\end{table}

\dataset differs from existing benchmarks by combining metric-coordinate answers with real outdoor service-robot data.
It contains repeated day and night visits to the same public street locations, paired robot and human viewpoints, and anonymized pedestrian scenes.
Among the listed benchmarks, only NaVQA and \dataset use a real robot for data collection, and only \dataset combines repeated visits with a dual-viewpoint protocol.
SpaceLocQA is marked as \emph{partial} outdoor because its sequences include outdoor campus walkways.
The distinguishing property of \dataset is therefore not outdoor recording alone, but the combination of repeated public-street visits and paired robot and human viewpoints, which directly supports evaluation of revisit handling, cross-domain robustness, and viewpoint mismatch.

\section{\dataset Collection Protocol}
\label{app:gangnam}

\paragraph{Route (anonymized).}
The fixed \dataset route starts and ends at a single office location, denoted by $E$, in a dense urban district of a large city. %
Each recording visits one of four nearby subway stations, denoted by $A$, $B$, $C$, and $D$, and returns to $E$ along the same path.
The one-way distance from $E$ to each station entrance ranges from approximately $400$ to $600\,\mathrm{m}$, producing round trips of roughly $0.9$-$1.2\,\mathrm{km}$.
The four destinations cover different urban conditions, including commercial street frontage, narrow side streets, and a wider boulevard segment.
This route design exposes the robot to a representative range of service-robot operating conditions.
% A (현재)
%The submission anonymizes the actual identities of $E$ and $A$-$D$ for double-blind review.
% B (수정)
The dataset anonymizes the actual identities of $E$ and $A$-$D$ for privacy.
%The submission anonymizes the actual identities of $E$ and $A$-$D$ for privacy.
Figure~\ref{fig:common_map} shows the common SLAM map reconstructed for this route by the onboard LiDAR-Inertial SLAM pipeline.
All eight recordings are registered to this single map, so that the office $E$, the four destinations $A$--$D$, and every day and night trajectory share one coordinate frame, which is what makes revisits of the same location directly comparable across conditions.
\begin{figure}[ht]
  \centering
  
  \begin{minipage}[t]{0.49\textwidth}\centering
    \vspace*{-40.5mm}
    \begin{subfigure}{\linewidth}
      \includegraphics[width=\linewidth]{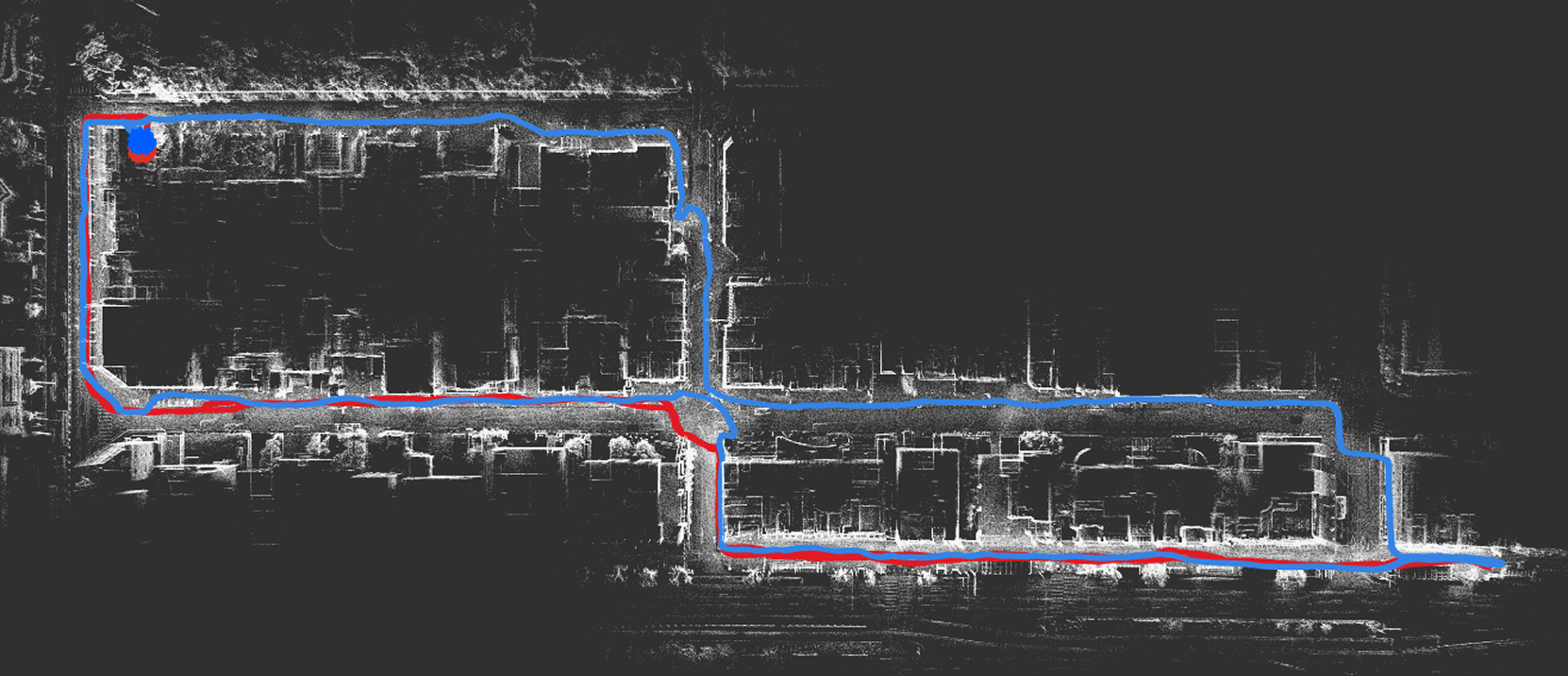}
      \caption{$E\!\to\!A$ ($R_1$ day, $R_8$ night).}\label{fig:gangnam}
    \end{subfigure}\\[3mm]
    \begin{subfigure}{\linewidth}
      \includegraphics[width=\linewidth]{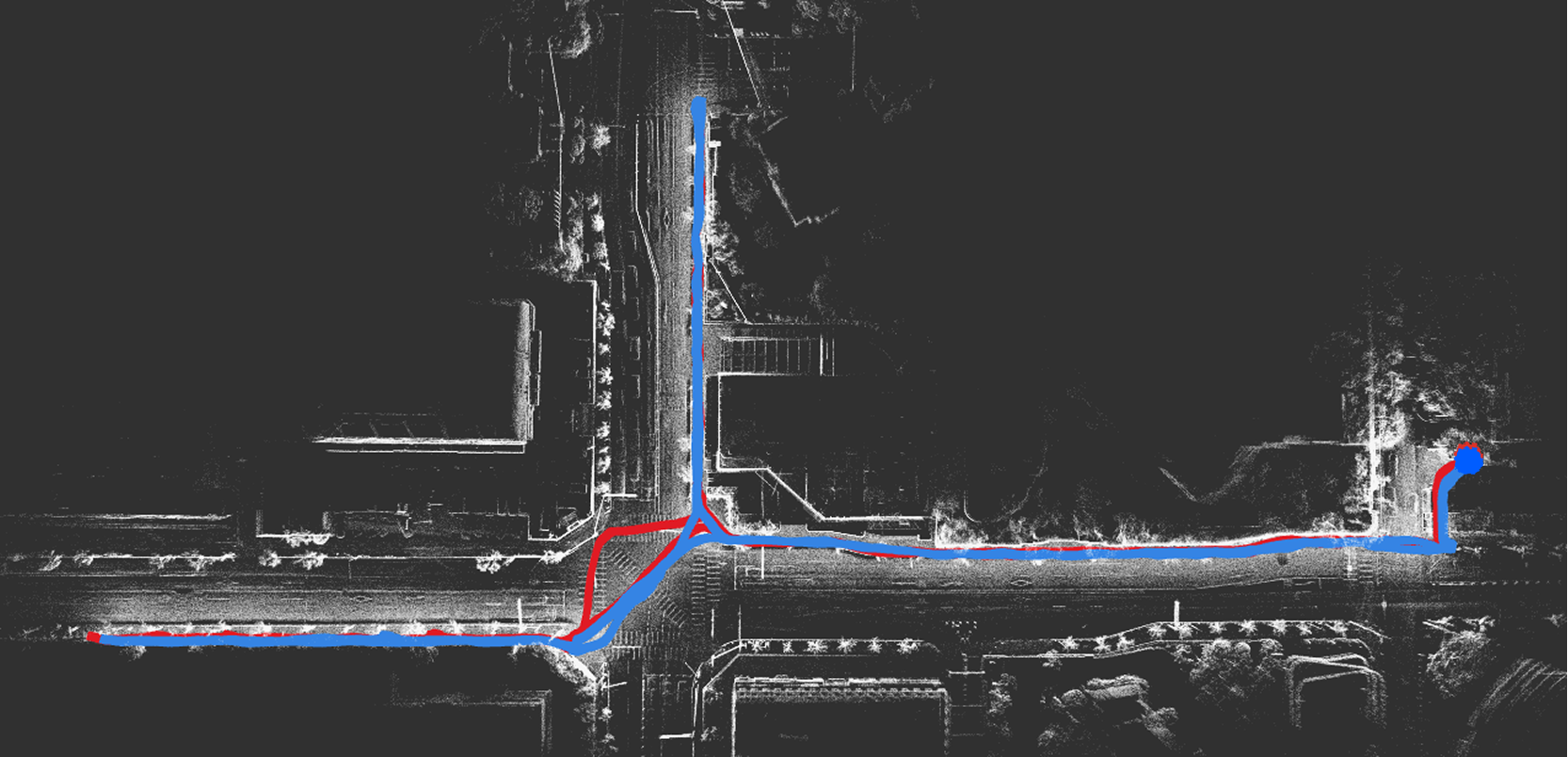}
      \caption{$E\!\to\!B$ ($R_2$ day, $R_7$ night).}\label{fig:seocho}
    \end{subfigure}
  \end{minipage}\hfill
  \begin{minipage}[c]{0.49\textwidth}\centering
    \setcounter{subfigure}{3}%
    \begin{subfigure}{\linewidth}
      \includegraphics[width=\linewidth]{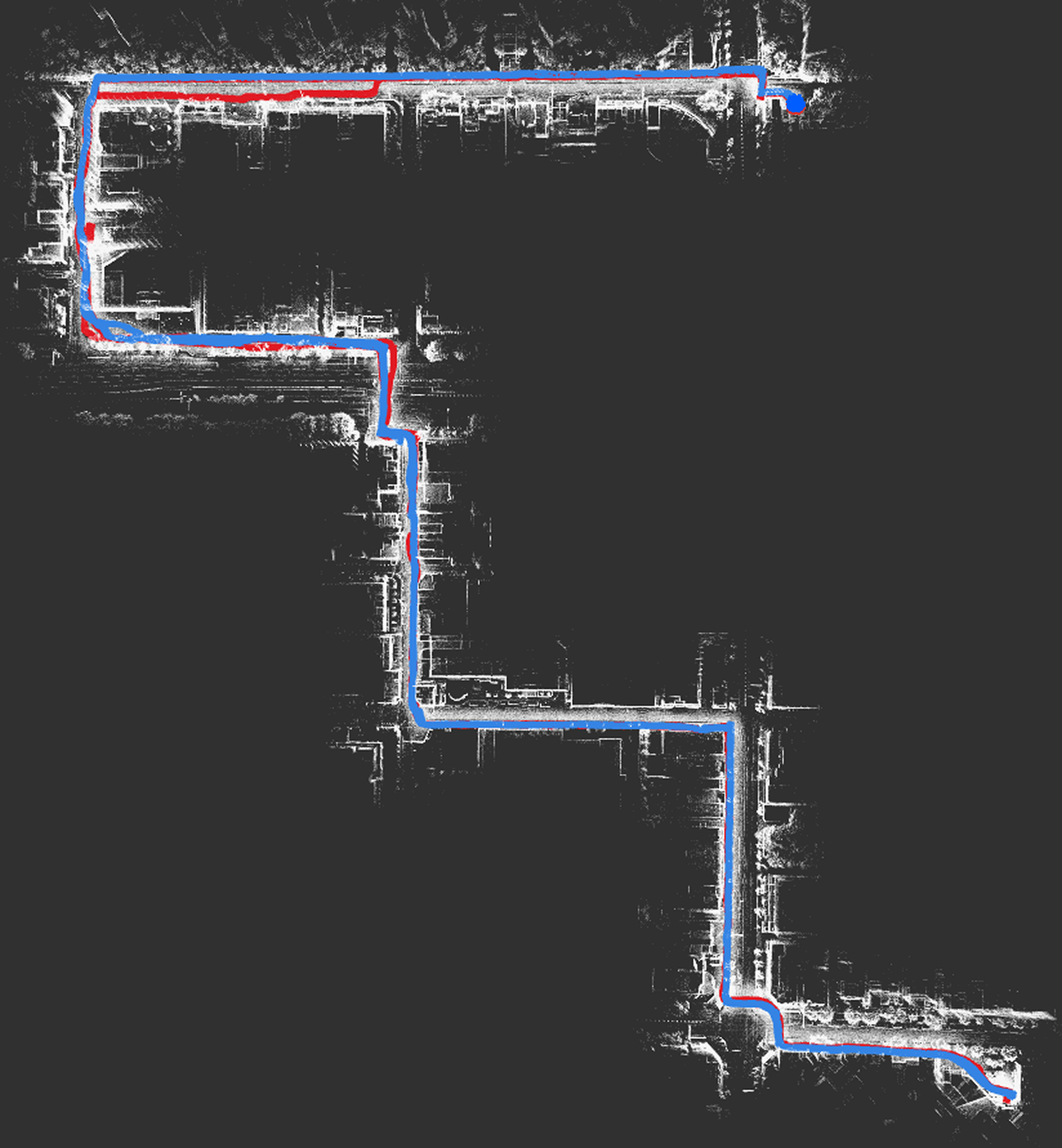}
      \caption{$E\!\to\!D$ ($R_5$ day, $R_4$ night).}\label{fig:yeoksam}
    \end{subfigure}
  \end{minipage}

  \vspace{3mm}
  \setcounter{subfigure}{2}%
  \begin{subfigure}{\linewidth}
    \includegraphics[width=\linewidth]{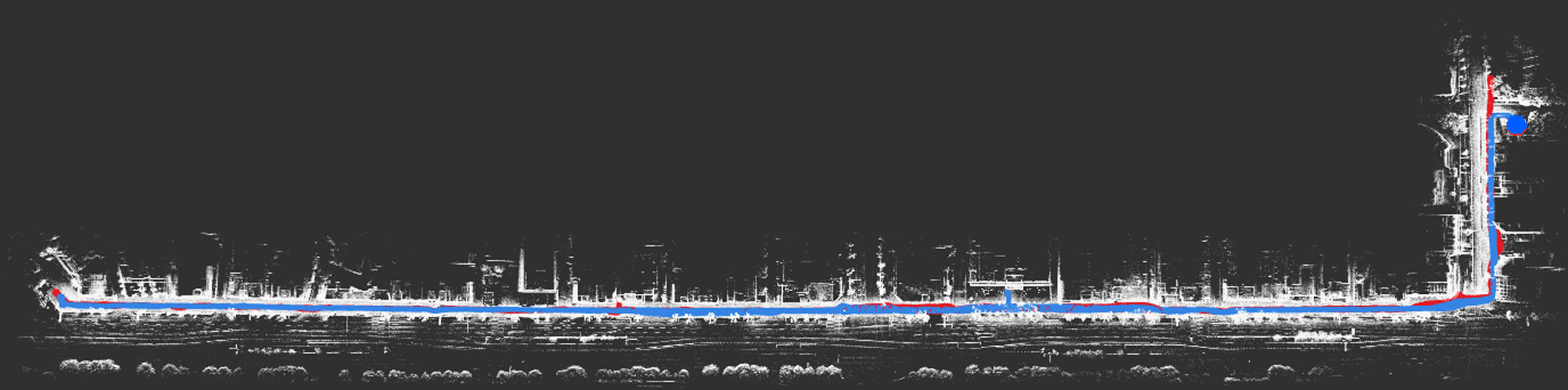}
    \caption{$E\!\to\!C$ ($R_6$ day, $R_3$ night).}\label{fig:yangjae}
  \end{subfigure}
  \caption{Day (red) and night (blue) trajectories for the four \dataset destinations, aligned on the common SLAM map.}
  \label{fig:gangnam-pairs}
\end{figure}

\begin{figure}[t]
\centering
\includegraphics[width=0.95\textwidth]{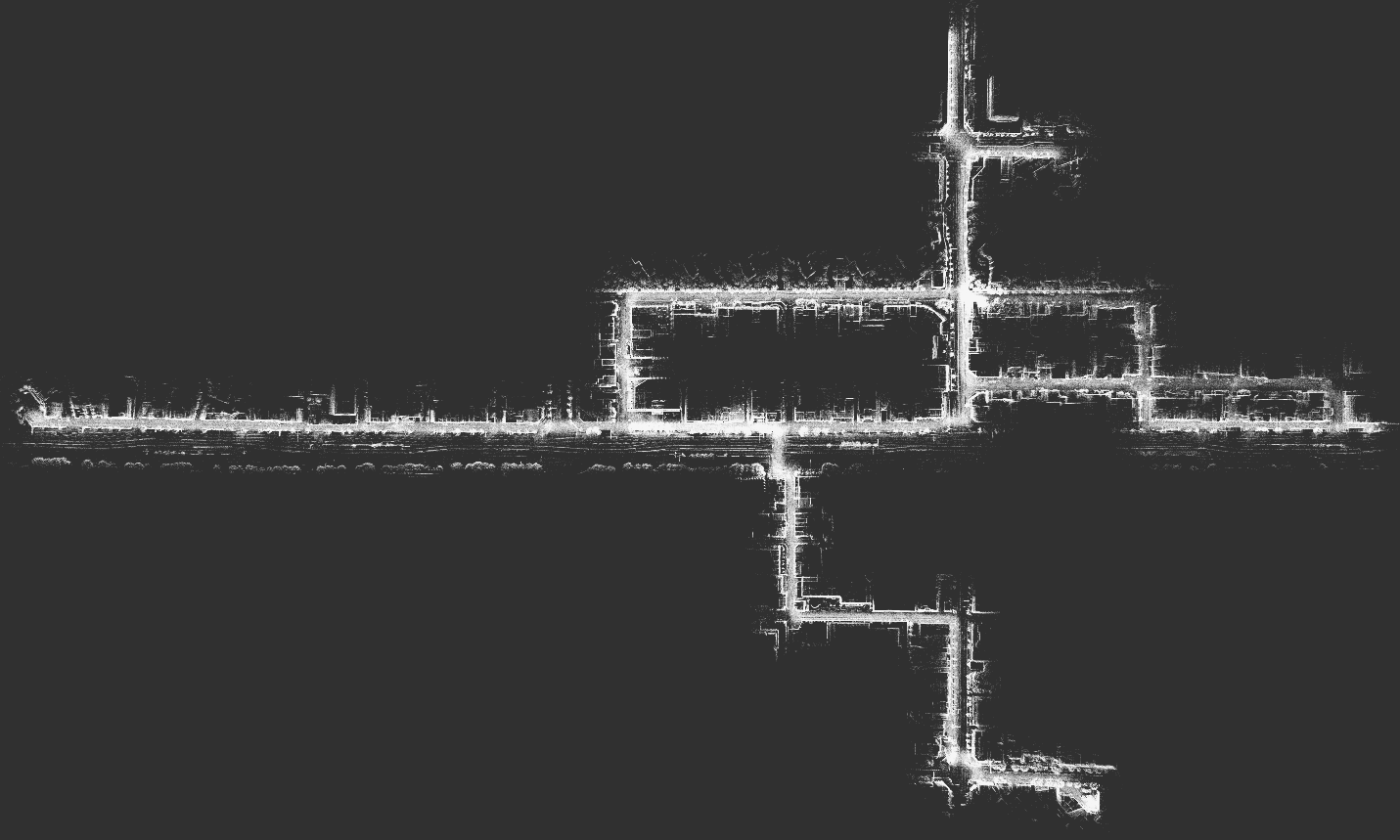}
\caption{The common SLAM map onto which all \dataset trajectories are aligned. The map is reconstructed by the onboard LiDAR-Inertial SLAM pipeline and shared across the eight day/night recordings, providing a single coordinate frame for the office ($E$) and the four station destinations ($A$--$D$). This shared frame is what makes revisits of the same location directly comparable across day, night, and route.}
\label{fig:common_map}
\end{figure}

\paragraph{Schedule and pairing.}
The collection schedule covers $2~\text{days} \times 4~\text{destinations} \times 2~\text{times of day}$.
Each destination has one daytime recording and one nighttime recording.
The recording pairs $(1,8)$, $(2,7)$, $(3,6)$, and $(4,5)$ share the same destination but differ in time of day.
These four pairs form the day and night cross-domain evaluation set for property (P2).
Table~\ref{tab:gangnam-schedule} summarizes the full schedule and indicates each pair in the ``Pair w/'' column.

\paragraph{Companion viewpoint.}
A human user walks alongside the quadruped with an RGB camera for a subset of recordings.
The system time synchronizes this human viewpoint with the robot's onboard sensors.
For owner viewpoint queries, the evaluation uses the robot pose at the matching timestamp as the ground truth coordinate.

\paragraph{Annotation procedure.}
The released labeling tool shows the synchronized RGB video and trajectory plot together.
Annotators browse the video, select the frame that corresponds to the query target, type the question, choose the SQA category, and save the annotation.
The selected frame provides the ground truth coordinate through its time-synchronized robot pose.
At least two independent annotators label each recording.
The annotation pipeline cross-validates the labels through relocalization and discards queries with more than $10\,\mathrm{m}$ inter-annotator disagreement.

\paragraph{Query budget.}
The benchmark targets approximately $360$ queries in total, balanced across basic, local, and global categories.
The query set also includes paired day-and-night examples for evaluating cross-domain robustness and dual viewpoint examples for evaluating viewpoint mismatch.

\paragraph{Pedestrian anonymization policy.}
\dataset contains real public-street recordings, so the raw videos inevitably include pedestrians and license plates unrelated to the project.
The release pipeline applies face and license-plate detection to every RGB frame and blurs detected regions with a Gaussian kernel.
The project releases only blurred frames and uses these anonymized frames for captioning, retrieval, and verification.
This policy ensures that all models in this paper, including the captioner, encoder, and verifier, operate on anonymized inputs.
It also prevents the reported results from relying on identifiable bystanders or private vehicle information.

\section{Per-Recording \dataset Results and Detailed Per-Trajectory Analyses}
\label{app:detailed-results}

Table~\ref{tab:gangnam-main} reports the per-recording success rates on all eight \dataset recordings at $\tau=15\,\mathrm{m}$.
The recording identifiers $R_1$-$R_8$ follow the schedule in Table~\ref{tab:gangnam-schedule}.
This comparison includes only open-source baselines, since no closed-source model result has been reported on \dataset.
\method consistently outperforms these baselines across recordings, confirming that the proposed pipeline transfers to outdoor low-viewpoint service robot data.

\begin{table}[ht]
    \centering\small
    \caption{Per-recording results on \dataset (success at $\tau=15\,\mathrm{m}$).}
    \label{tab:gangnam-main}
    \begin{tabular}{llccccccccc}
      \toprule
      Method & Backbone & $R_1$ & $R_2$ & $R_3$ & $R_4$ & $R_5$ & $R_6$ & $R_7$ & $R_8$ & Overall \\
      \midrule
      Meta-Memory~\citep{mao2025metamemory} & open-source$^\dagger$
         & 15.6 & 31.1 & 0.0 & 6.7 & 20.0 & 8.9 & 26.7 & 15.6 & 15.6 \\
      ReMEmbR~\citep{anwar2024remembr} & open-source$^\dagger$
         & 24.4 & 33.3 & 4.4 & 8.9 & 24.4 & 11.1 & 31.1 & 6.7 & 18.0 \\
      \textbf{\method (ours)} & open-source
         & \textbf{55.6} & \textbf{60.0} & \textbf{24.4} & \textbf{40.0} & \textbf{60.0} &
  \textbf{31.1} & \textbf{57.8} & \textbf{33.3} & \textbf{45.3} \\
      \bottomrule
    \end{tabular}
    \\[2pt]
    {\footnotesize $^\dagger$ Reproduced by us under the fair constraint (Qwen2.5-32B-AWQ agent and Qwen2.5-7B
  verifier~\cite{bai2025qwen25vl}).}
  \end{table}

\paragraph{Per-recording analysis.} 
The success rate on \dataset varies substantially across recordings, ranging from $24.4\%$ to $60.0\,\%$.
Daytime recordings generally achieve higher accuracy than their night counterparts.
Among the four-day and night pairs, the daytime recording performs better in three pairs: $R_1{=}55.6$ vs.\ $R_8{=}33.3$, $R_5{=}60.0$ vs.\ $R_4{=}40.0$, and $R_2{=}60.0$ vs.\ $R_7{=}57.8$.
This trend reflects the illumination domain shift that \dataset is designed to evaluate.
Route length also affects performance.
The longest routes, $R_3$ and $R_6$, correspond to the destination-$C$ round-trip and contain the largest number of segments.
These recordings yield the lowest success rates, $24.4\,\%$ and $31.1\,\%$, because longer trajectories introduce more same-class distractor landmarks and require Binary Tracking to search over longer anchor intervals.

\begin{table}[ht]
      \centering\small
      \caption{Per-trajectory success rates on SpaceLocQA (45 queries each, success at $\tau=15\,\mathrm{m}$).}
      \label{tab:per-traj}                                                                                
      \begin{tabular}{rlcccccc}                      
        \toprule                                                                                          
           & Method & 0 & 1 & 2 & 3 & 4 & 5 \\       
        \midrule                                                                                          
        1 & ReMEmbR$^\dagger$~\citep{anwar2024remembr}      & 51.1 & 48.9 & 62.2 & 57.8 & 42.2 & 40.0 \\
        2 & Meta-Memory$^\dagger$~\citep{mao2025metamemory} & 37.8 & 48.9 & 48.9 & 42.2 & 37.8 & 51.1 \\
        \textbf{3} & \textbf{\method (ours)}                & \textbf{75.6} & \textbf{57.8} & \textbf{71.1} & \textbf{68.9} & \textbf{62.2} & \textbf{68.9} \\
        \bottomrule                                                                                       
      \end{tabular}                                                                                       
      \\[2pt]
      {\footnotesize $^\dagger$ Reproduced by us under the open-source                                    
      constraint (Qwen2.5-32B-AWQ agent and Qwen2.5-7B verifier).}
\end{table}  

\paragraph{Per-trajectory SpaceLocQA results.}
Table~\ref{tab:per-traj} reports the per-trajectory success rates of \method on the six SpaceLocQA trajectories at $\tau=15\,\mathrm{m}$.
Each trajectory contains 45 queries.
The comparison uses the v2 reproduction numbers under the fair open-source constraint, with the same agent and verifier based on Qwen2.5-32B-AWQ~\cite{bai2025qwen25vl}. 
Trajectory~1 is the hardest case, as it covers a kilometer-scale campus route with a geographically separated dormitory cluster.
The remaining trajectories are more compact and yield consistently higher accuracy.

\paragraph{Mean and median errors.}
On the 270-query SpaceLocQA benchmark, \method achieves a mean localization error of $59.7\,\mathrm{m}$ and a median error of $4.0\,\mathrm{m}$.
The large gap between the mean and median indicates that most predictions are close to the target, while a small number of catastrophic failures dominate the mean error.
These failures mainly occur on the hardest trajectory, as discussed in Appendix~\ref{app:ablation-failure}.

\paragraph{Success-rate curve.}
Table~\ref{tab:tau-curve} reports the success rate of \method under different metric tolerances $\tau$.
This curve shows how performance changes as the localization threshold becomes stricter or more relaxed.

 \begin{table}[ht]
    \centering\small
    \caption{Success rate as a function of metric tolerance $\tau$ on SpaceLocQA ($270$ queries),
    under the fair open-source constraint (Qwen2.5-$32$B-AWQ agent + Qwen2.5-VL-$7$B verifier).
    Meta-Memory and ReMEmbR are our faithful reproductions.}
    \label{tab:tau-curve}
    \begin{tabular}{lcccc}
      \toprule
      Method (SR \% $\uparrow$)             & $\tau{=}5$\,m & $\tau{=}10$\,m & $\tau{=}15$\,m & $\tau{=}20$\,m \\
      \midrule
      Meta-Memory~\citep{mao2025metamemory} & 32.2 & 41.9 & 44.6 & 47.8 \\
      ReMEmbR~\citep{anwar2024remembr}      & 41.1 & 46.7 & 50.1 & 54.1 \\
      \textbf{\method (ours)}               & \textbf{54.1} & \textbf{64.1} & \textbf{67.4} & \textbf{69.3} \\
      \bottomrule
    \end{tabular}
  \end{table}

\paragraph{Tool-call frequency.}
The tool-call statistics come from the logged agent traces on the \dataset runs.
Under the headline 32B-agent configuration, \method issues an average of $3.0$ tool calls per query.
This includes one \textsc{final} action and, on average, two retrieval or Binary Tracking actions.
Semantic-similarity retrieval accounts for the largest share of actions ($37\%$), followed by the \textsc{final} action ($27\%$), spatial-range retrieval ($27\%$), Binary Tracking ($5\%$), and the legacy graph-based memory-integration tool ($4\%$).
Binary Tracking accounts for only a small fraction of all actions because the agent invokes it mainly for route-constrained global queries rather than at every step.
Nevertheless, it plays the decisive role for that category, as discussed in Section~\ref{sec:bs}.
The agent reaches an explicit \textsc{final} answer within its step budget on $81\%$ of queries.
For the remaining queries, the lock-on-Y fallback in Appendix~\ref{app:hardening} supplies the predicted coordinate.

\paragraph{Per-step latencies.}
Under the configuration in Appendix~\ref{app:implementation}, a single semantic retrieval against the local vector database takes approximately $30\,\mathrm{ms}$.
A single verifier call is much more expensive and takes on the order of one second per candidate image.
Since a query may verify several candidates across the three views, verifier calls dominate end-to-end latency.
A complete query resolves in roughly $45$--$70\,\mathrm{s}$ on average.
The average latency is $59\,\mathrm{s}$ on SpaceLocQA, $71\,\mathrm{s}$ under the 32B-agent \dataset configuration, and $44\,\mathrm{s}$ with the 7B agent.
These results support the motivation for Binary Tracking, which reduces expensive verifier calls by pruning the trajectory interval before verification.
Further latency reductions through verifier batching or distillation remain natural engineering directions.

\paragraph{Latency comparison with prior open-source pipelines.}
Table~\ref{tab:latency} reports end-to-end per-query latency on SpaceLocQA under the fair open-source constraint.
All systems use the same Qwen2.5-32B-AWQ planning agent and Qwen2.5-VL-7B verifier.
The reported speedup divides each baseline's latency by the latency of \method.
The results substantiate the ${>}1.5\times$ speedup reported in the abstract and Section~\ref{sec:intro}.
This speedup comes from the structure of the inference process.
\method invokes the expensive vision-language verifier mainly once per query, after Binary Tracking reduces the search space to the final leaf interval.
In contrast, graph-based baselines often re-verify many candidates across multiple agent steps, which increases end-to-end latency.

 \begin{table}[ht]
      \centering\small
      \setlength{\tabcolsep}{6pt}
      \caption{End-to-end per-query latency on SpaceLocQA ($270$ queries) under the fair open-source constraint.}
      \label{tab:latency}
      \begin{tabular}{lcc}
        \toprule
        Method (open-source) & Latency / query (s) $\downarrow$ & Speedup vs.\ \method \\
        \midrule
        Meta-Memory~\citep{mao2025metamemory} & $111.6$ & $1.89\times$ \\
        ReMEmbR~\citep{anwar2024remembr}      & $120.3$ & $2.04\times$ \\
        \textbf{\method (ours)}               & $59$    & -- \\
        \bottomrule
      \end{tabular}
  \end{table}

\paragraph{Low-VRAM reproducibility.}
Although the main experiments use two NVIDIA RTX 6000 Ada Generation GPUs, \method can run in a lighter configuration with only 7B-class open-source models.
This configuration replaces the captioner, text encoder, planning agent, and visual verifier with 7B-scale open-source counterparts while preserving the Binary Tracking primitive.
The full inference pipeline fits within roughly $20\,\mathrm{GB}$ of GPU VRAM, making \method feasible on a single consumer GPU and more practical for resource-constrained service robots.

\begin{table}[ht]
  \centering\small
  \setlength{\tabcolsep}{5pt}
  \caption{Effect of the planning-agent size on \dataset (all 360
  queries, success at $\tau=15\,\mathrm{m}$).}
  \label{tab:agent-size}
  \begin{tabular}{lccccrr}
    \toprule
    Planning agent & Basic ↑ & Local ↑ & Global ↑ & Overall ↑ & Steps/q $\downarrow$& Latency/q $\downarrow$ \\
    \midrule
    32B-AWQ (headline)      & 50.0 & 43.3 & \textbf{42.5} & \textbf{45.3} & 2.96 & $71\,\mathrm{s}$ \\
    7B (low-VRAM)           & \textbf{55.8} & \textbf{45.8} & 30.0 & 43.9 & 2.92 & $\mathbf{44\,\mathrm{s}}$ \\
    \midrule
    $\Delta$ (7B $-$ 32B)   & $+5.8$ & $+2.5$ & $-12.5$ & $-1.4$ & $-0.04$ & $-27\,\mathrm{s}$ \\
    \bottomrule
  \end{tabular}
\end{table}

Table~\ref{tab:agent-size} compares this low-VRAM configuration with the headline 32B-agent configuration on all 360 \dataset queries.
The 7B-only configuration loses only $1.4\,\%$ overall accuracy and achieves a ${\sim}1.6\times$ speedup, reducing per-query latency from $71\,\mathrm{s}$ to $44\,\mathrm{s}$.
It performs competitively on basic and local queries, while the main deficit appears in the global category ($-12.5\,\%$), where the larger agent more reliably decomposes route-constrained queries into the two anchors required by Binary Tracking.
These results show that the low-VRAM version of \method remains close to the headline configuration while substantially reducing hardware and latency requirements.

\section{Ablation Study and Failure Analysis}
\label{app:ablation-failure}

\subsection{Extended Ablation Study}
\label{app:ablation}

Table~\ref{tab:ablation} reports the extended ablation on SpaceLocQA.
The study evaluates the contribution of four components: multi-view memory (MV), Binary Tracking (BT), lock-on-Y rerank (LY), and anchor-aware verification pool (AA).
The first block removes each component individually, while the second block removes pairwise combinations.
Each configuration disables the corresponding component or components from the released pipeline.

\begin{table}[h]
  \centering\small
  \caption{Extended ablation on SpaceLocQA (270 queries, success at $\tau=15\,\mathrm{m}$).}
  \label{tab:ablation}
  \begin{tabular}{lcccc}
    \toprule
    Configuration                                & Basic ↑ & Local ↑ & Global ↑ & Overall ↑ \\
    \midrule
    \textbf{\method (full)}                      & \textbf{74.4} & \textbf{65.6} & \textbf{62.2} & \textbf{67.4} \\
    \midrule
    w/o BT                                        & 74.4 & 63.3 & 52.2 & 63.3 \\
    w/o MV                                        & 76.7 & 63.3 & 48.9 & 63.0 \\
    w/o LY                                        & 74.4 & 65.6 & 55.6 & 65.2 \\
    w/o AA                                        & 74.4 & 65.6 & 56.7 & 65.2 \\
    \midrule
    w/o BT \& LY                                  & 74.4 & 64.4 & 47.8 & 62.2 \\
    w/o BT \& AA                                  & 74.4 & 65.6 & 50.0 & 63.0 \\
    w/o MV \& BT                                  & 76.7 & 60.0 & 42.2 & 59.6 \\
    \bottomrule
  \end{tabular}
\end{table}

The two core components, Binary Tracking and multi-view memory, produce the largest drops on the global category when removed.
Removing Binary Tracking reduces global accuracy by $10.0\,\%$, and removing multi-view memory reduces it by $13.3\,\%$.
These results confirm that the two components directly support the path-constrained reasoning regime that motivates the proposed primitive.
The two supplementary mechanisms, LY and AA, play a smaller but complementary role.
Removing each of them individually reduces overall accuracy by about $2\,\%$, showing that they provide modest but consistent robustness gains.

\subsection{Failure Analysis}
\label{app:failure-analysis}

\paragraph{Failure categorization.}
This analysis examines the $197$ failed queries out of the $360$ \dataset queries under the headline 32B-agent configuration.
A failure occurs when \method predicts a coordinate more than $15\,\mathrm{m}$ away from the ground truth.
The analysis classifies each failure into one of three categories, as shown in Table~\ref{tab:fail-cats}.
The categorization uses logged agent traces, including the action sequence, verifier votes, and whether the agent invoked Binary Tracking.
The released code includes the heuristic used for this categorization, making the analysis fully reproducible.
Representative cases also undergo manual spot checks.
The largest category consists of \emph{agent reasoning errors}.
These failures occur when the agent emits a \textsc{final} answer before invoking Binary Tracking on a clearly route-constrained query, or when it exhausts its step budget.
Only $49$ of the $120$ global queries trigger Binary Tracking, which suggests that improving the agent's tool selection is the largest opportunity for improving \dataset accuracy.
The next category consists of \emph{verifier and retrieval false negatives}.
These failures occur when the 7B-class verifier rejects or mis-ranks an image that actually contains the queried entity.
They often involve small objects, partial occlusion, or non-English street signage, especially Korean text.
Genuine memory misses, where no segment view captures the target, remain rare.
This analysis folds such cases into the same category because separating them would require per-segment ground-truth matching.
The last category consists of \emph{anchor disambiguation failures}.
These failures occur when the agent invokes Binary Tracking but commits to the wrong same-class landmark, such as one of several convenience stores, even with the lock-on-Y rerank.
The three categories are computed directly from the logged traces and sum exactly to the $197$ failures.

\begin{table}[ht]
  \centering\small
  \caption{Failure mode categorization on the $197$ \dataset
  queries with $\hat{p}$-error~$>15\,\mathrm{m}$ under the headline
  32B-agent configuration.}
  \label{tab:fail-cats}
  \begin{tabular}{lcc}
    \toprule
    Category                              & Count & Share \\
    \midrule
    Agent reasoning errors                & 84 & 42.6\,\% \\
    Verifier / retrieval false negatives  & 83 & 42.1\,\% \\
    Anchor disambiguation failures        & 30 & 15.2\,\% \\
    \bottomrule
  \end{tabular}
\end{table}

\paragraph{Representative examples.}
Agent reasoning errors include route-constrained queries such as
\textit{``Where is the nearest bus stop?''} and
\textit{``Where is the nearest JAJU sign?''}.
In these cases, the agent performed a single semantic retrieval and then issued
\textsc{final} without invoking Binary Tracking.
Anchor disambiguation failures include queries such as
\textit{``Where is the smoking area on the way from the CU convenience store to KFC?''}
and
\textit{``Where is the blue bicycle on my way from the KFC to the PC room?''}.
In these cases, Binary Tracking ran, but the verifier selected the wrong
same-class landmark.
Verifier false negatives include appearance-based queries such as
\textit{``I see a large gray electrical utility box on the sidewalk and a sign reading `C' on a building\ldots''}.
Such scenes are visually generic, and the 7B verifier often fails to
disambiguate the correct segment.

\paragraph{Plausible mitigations.}
Stricter prompting can reduce the dominant agent-reasoning errors by forcing
the agent to enumerate all named entities in the query before choosing a tool.
The prompt can also require Binary Tracking whenever the query contains two
route-defining anchors.
This change provides the cheapest and highest-leverage fix because fewer than half of the global queries currently trigger Binary Tracking.
A stronger verifier, such as a 32B-class open-source VLM, could reduce many
verifier false negatives, especially for small objects, occlusions, and
non-English signage.
The current experiments do not use such a verifier because of the GPU memory
constraints on the two-GPU workstation.
Anchor disambiguation failures may decrease by extending the lock-on-Y rerank
to handle three-anchor queries explicitly.

\section{Correctness Condition and Empirical Check}
\label{app:correctness}
This appendix recalls the sufficient condition under which Binary Tracking reaches a leaf interval containing a true target segment.
We first state the condition assuming a unique target segment $i^\star$, and then relax it to the multi-target case that arises in \dataset.

Because a route-constrained query restricts the search to the directional sub-path between the anchor segments $i_X$ and $i_Y$, the target $Z$ typically appears at a single segment within this interval, even when the full round-trip recording revisits the location.
For each visited interval $[\ell,r]$ with midpoint $m=\lfloor(\ell+r)/2\rfloor$, let $H^\star(\ell,m,r)$ denote the half that contains $i^\star$, and let $H^-(\ell,m,r)$ denote the other half.
Let $\sigma(i)$ denote the cosine similarity between the target query embedding and the caption embedding of segment $S_i$, as used in the evidence score in Section~\ref{sec:bs}.
A sufficient condition for Binary Tracking to keep the correct half at every split is
\begin{equation}
    \max_{i \in H^\star(\ell,m,r)} \sigma(i)
    >
    \max_{i \in H^-(\ell,m,r)} \sigma(i).
    \label{eq:correctness}
\end{equation}
This condition does not require the true target segment $i^\star$ itself to have the highest semantic score over the entire interval; it only requires the best candidate on the correct side to score higher than the best candidate on the wrong side at each split.
The verifier then resolves the remaining ambiguity within the final leaf interval and selects the predicted segment $\hat{i}$.

\paragraph{Multiple target segments under revisits.}
In the less common case where the anchor interval spans the turnaround point of a round trip, the same entity $Z$ can correspond to more than one segment at the segment level.
We therefore define the set of acceptable target segments within the anchor interval $[i_X, i_Y]$,
\begin{equation}
  \mathcal{I}^\star = \bigl\{\, i \in [i_X, i_Y] : \lVert \mathbf{p}_i - \mathbf{p}^\star \rVert_2 < \tau \,\bigr\},
\end{equation}
collecting every segment whose representative pose lies within the success tolerance $\tau$ of the ground-truth coordinate $\mathbf{p}^\star$.
Because revisits of the same landmark occur at nearly the same physical location, their representative poses typically fall within $\tau$ of one another, so all such segments are equally acceptable under the success metric of Section~\ref{sec:problem}.
Equation~\ref{eq:correctness} then needs to hold only at splits where exactly one half contains a member of $\mathcal{I}^\star$: letting $H^\star$ be that half, it guarantees that Binary Tracking descends toward an acceptable target.
When both halves contain a member of $\mathcal{I}^\star$, either descent reaches an acceptable target, so no condition is required at that split.

The multi-view memory in Section~\ref{sec:memory} helps satisfy this condition in practice.
A target segment may have a weak score in one view but a strong score in another view, increasing the chance that the correct half wins the split.

\section{Naive Cross-Trajectory Retrieval Experiment}
\label{app:cross-traj}

\paragraph{Setup.}
\method builds a separate memory for each trajectory, as prior systems do, and applies a per-trajectory filter to every retrieval call (Appendix~\ref{app:agent-tools}).
This appendix tests whether repeated visits to the same locations in \dataset can help retrieval with a simple cross-trajectory relaxation.
The naive cross-trajectory mode removes the per-trajectory filter at retrieval time.
Under this setting, \textsc{SSR}, \textsc{SRR}, and the internal retrieval inside \textsc{BT} can return candidate segments from all available trajectories of the same route.
The rest of the \method pipeline remains unchanged, including multi-view memory, the hardening mechanisms, and the verifier call on the leaf interval.
This experiment requires no retraining, re-indexing, or algorithmic change.
It only changes the partition tag passed to the vector database.

\paragraph{Result.}
The experiment compares the headline single-trajectory configuration with the naive cross-trajectory mode on the four-day and night pairs of \dataset.
Each query can retrieve candidates from both recordings in its day-and-night pair.
The naive cross-trajectory mode produces no notable gain over the single-trajectory configuration.
It yields a small improvement on basic queries but a small drop on global queries, and these effects approximately cancel in the overall accuracy.
Because this probe does not change the headline conclusion, this appendix reports the result qualitatively rather than adding a separate results table.
A principled cross-trajectory integration method remains future work.

\paragraph{Discussion.}
Naive cross-trajectory retrieval does not benefit much from the union of trajectories for two structural reasons.
A basic or local query already retrieves the correct segment from its own trajectory in most cases, so additional candidates from other trajectories mainly crowd the verifier pool.
A global query also imposes a route constraint, and segments from different trajectories usually lie outside the anchor interval defined by the current query.
Such candidates, therefore, add distractors rather than improving the route-consistent answer.
A more principled cross-trajectory method would need to align segments across trajectories before retrieval and then run Binary Tracking on the aligned representation.
This direction remains future work.

\paragraph{Offline memory-build cost on new routes.}
The captioning step takes several GPU-hours on the 68-minute SpaceLocQA corpus and roughly one hour on a single \dataset recording.
Binary Tracking substantially reduces retrieval-time cost, but a newly visited route still requires offline or background captioning and embedding before the system can answer queries.
This memory-build requirement remains a practical barrier to fully online deployment when the robot encounters an unseen route.
Reducing the latency of memory construction is an important direction for future work.

\section{Glossary and Detailed Task Definitions}
\label{app:glossary}

This appendix provides extended definitions of the terms used in the main paper.
Readers familiar with spatial question answering for embodied agents may skip this section.

\paragraph{Spatial question answering (SQA).}
SQA follows the task formulation of Meta-Memory~\citep{mao2025metamemory} and ReMEmbR~\citep{anwar2024remembr}.
Given a robot's egocentric trajectory, consisting of images and 2D poses, and a natural-language question, the system outputs a metric 2D coordinate $\hat{\mathbf{p}}$ as the answer.
The evaluation counts a prediction as successful when its Euclidean distance to the annotated ground-truth coordinate falls below a fixed tolerance $\tau$.
This work adopts $\tau = 15\,\mathrm{m}$ throughout, following prior work.
SQA differs from embodied question answering~\citep{das2018eqa,majumdar2024openeqa} because the agent does not navigate during inference.
It also differs from episodic memory QA~\citep{datta2022emqa} because the answer takes the form of a metric coordinate rather than a category label or a natural-language response.

\paragraph{Query categories.}
SpaceLocQA~\citep{mao2025metamemory} groups questions into three categories of increasing difficulty.

\begin{itemize}
    \item \textbf{Basic queries} require recalling a single object or landmark.
    Examples include \textit{``Where is a dry cleaner?''} and \textit{``Find a vending machine.''}
    These queries usually require semantic retrieval followed by visual verification.
    \item \textbf{Local queries} require integrating multiple attributes or objects within a small region.
    Examples include \textit{``Which room contains a refrigerator, a microwave, and a window?''} and \textit{``Find the corner with both a coffee shop and a bus stop.''}
    These queries usually combine semantic retrieval with spatial-range retrieval over a local neighborhood.
    \item \textbf{Global queries} require reasoning over spatially separated entities along a long trajectory.
    Examples include \textit{``Where is the vending machine on the route from the basketball court to the football field?''} and \textit{``Find the AED nearest the lakeside.''}
    These queries often involve route constraints or anchor-relative reasoning, and they motivate Binary Tracking in this work.
\end{itemize}

\paragraph{Retrieval primitives (SSR, SRR, MI).}
This work retains the three retrieval primitive names introduced by \citet{mao2025metamemory}.
\emph{Semantic-similarity retrieval} (\textsc{SSR}) returns the top-$k$ segments ranked by cosine similarity between the query embedding and each segment caption embedding.
\emph{Spatial-range retrieval} (\textsc{SRR}) returns segments whose poses lie within a given radius of a query point.
\emph{Memory-integration} (\textsc{MI}) constructs a topological waypoint graph from retrieved landmark positions and runs Dijkstra's algorithm~\citep{dijkstra1959note} to produce a path between two query points.
Binary Tracking (\textsc{BT}) is the new retrieval primitive introduced in this paper, as described in Section~\ref{sec:bs}.

\paragraph{Memory segment.}
Following \citet{anwar2024remembr} and \citet{mao2025metamemory}, the system divides the robot trajectory into fixed-length time segments.
The released configuration uses $\Delta t = 1.5\,\mathrm{s}$, as described in Appendix~\ref{app:implementation}.
Each segment contains four evenly spaced frames, their $2{\times}2$ concatenated image, one or more captions, a mean 2D pose, and a unique segment index.
The robot's memory for one trajectory consists of the full ordered set of these segments.

\paragraph{Verifier ensemble.}
Retrieval returns a small set of candidate segments.
A vision-language model then judges whether the candidate images contain the queried entity.
This work refers to one such judgment as a \emph{verifier call}.
It refers to a batch of judgments across multiple views or related candidates as a \emph{verifier ensemble}.
Section~\ref{sec:agent} and Appendix~\ref{app:hardening} describe how the agent uses these verifier calls.

\paragraph{Service robot deployment.}
This phrase refers to a deployment scenario in which a robot accompanies a human user along a familiar, repeated route, such as a quadruped walking with a person to the same office or store.
This scenario motivates three properties of \dataset, labeled (P1)--(P3) throughout Appendix~\ref{app:gangnam}.
(P1) captures repeated visits to the same locations.
(P2) captures domain shifts across visits, such as day versus night or weather changes.
(P3) captures the viewpoint mismatch between a low-mounted robot camera and a human user's head-mounted viewpoint, as discussed in Section~\ref{sec:dataset}.

\section{Background on Foundation Models and Retrieval Infrastructure}
\label{app:background}

\paragraph{Open-source language and vision-language models.}
The pipeline uses an open-source instruction-tuned LLM as the planning agent~\citep{yang2024qwen25} and deploys it in quantized form~\citep{lin2024awq}.
The captioner and visual verifier share a single open-source video VLM~\citep{bai2025qwen25vl}.
All models run locally on a single workstation and require no external API access.
Their permissive licenses also support reproducible open-source deployment.

\paragraph{Embedding and retrieval infrastructure.}
The memory system embeds captions with an open-source $1024$-dimensional text encoder~\citep{lee2024mxbai} and indexes them in a local vector database~\citep{wang2021milvus}.
This design follows the broader retrieval-augmented generation paradigm~\citep{lewis2020rag}, where the system retrieves relevant memory entries before reasoning over them.
The implementation uses these components to match prior work~\citep{mao2025metamemory} and isolate the contribution of Binary Tracking.
The Binary Tracking algorithm itself does not depend on a specific captioner, encoder, or vector database.

\fi
\ifshowbody\else
\bibliography{references}   %
\fi
\end{document}